%% file: main.tex
\documentclass{clv3}
\usepackage{times,latexsym}
\usepackage{url}
\usepackage[T1]{fontenc}
\usepackage{dsfont}
\usepackage{mathtools}
\usepackage{booktabs}
\usepackage{framed}
\usepackage{varwidth}
\usepackage{array}
\usepackage{hyperref}
\usepackage{xcolor}
\usepackage{amsmath,amsfonts,amssymb,verbatim,ifthen,amsthm}
\usepackage{color}
\usepackage{multirow}
\usepackage[all]{xy}
\usepackage{setspace}
\usepackage{graphicx}
\usepackage{nameref}
\usepackage{bbm}
\usepackage{csquotes}
\usepackage[leftmargin=3em]{quoting}
\usepackage{subcaption}
\usepackage{placeins}
\usepackage{lscape} 
\usepackage{enumitem}
\usepackage{arydshln}
\usepackage{tabularx}
\usepackage{multirow}
\usepackage{xcolor}
\usepackage[title]{appendix}
\usepackage{soul}

\DeclareMathOperator*{\argmax}{arg\,max}

\newtheorem{definition}{Definition}
\newcommand{\com}[1]{}

\begin{document}

\runningtitle{CausaLM}
\runningauthor{Feder, Oved, Shalit and Reichart}

\title{CausaLM: Causal Model Explanation Through Counterfactual Language Models}

\author{Amir Feder \\
  {\tt feder@campus.technion.ac.il} \\
  Nadav Oved \\
  {\tt nadavo@campus.technion.ac.il} \\
  Uri Shalit \\
  {\tt urishalit@technion.ac.il} \\
  Roi Reichart \\
  {\tt roiri@technion.ac.il} \\}


\maketitle

\begin{abstract}
\input{abstract}

\end{abstract}

\input{chapters/intro.tex}

\input{chapters/prev.tex}
\input{chapters/causal.tex}
\input{chapters/data.tex}

\input{chapters/tasks.tex}
\input{chapters/results.tex}
\input{chapters/discussion.tex}

\section*{Acknowledgements}
This research was supported by the Zuckerman Fund to the Technion Artificial Intelligence Hub (Tech.AI).

\newpage

\begin{appendices}
\input{chapters/appendix.tex}

\end{appendices}

\newpage

\bibliographystyle{compling}
\starttwocolumn
\bibliography{bibliography}

\end{document}

%% file: abstract.tex
Understanding predictions made by deep neural networks is notoriously difficult, but also crucial to their dissemination. As all machine learning based methods, they are as good as their training data, and can also capture unwanted biases. While there are tools that can help understand whether such biases exist, they do not distinguish between correlation and causation, and might be ill-suited for text-based models and for reasoning about high level language concepts. A key problem of estimating the causal effect of a concept of interest on a given model is that this estimation requires the generation of counterfactual examples, which is challenging with existing generation technology. To bridge that gap, we propose CausaLM, a framework for producing causal model explanations using counterfactual language representation models. Our approach is based on fine-tuning of deep contextualized embedding models with auxiliary adversarial tasks derived from the causal graph of the problem. Concretely, we show that by carefully choosing auxiliary adversarial pre-training tasks, language representation models such as BERT can effectively learn a counterfactual representation for a given concept of interest, and be used to estimate its true causal effect on model performance. A byproduct of our method is a language representation model that is unaffected by the tested concept, which can be useful in mitigating unwanted bias ingrained in the data.~\footnote{Our code and data are available at: \url{https://amirfeder.github.io/CausaLM/}. Accepted for publication in \textit{Computational Linguistics} journal: 4 March 2021.}

%% file: chapters/intro.tex
\section{Introduction}
\label{sec:intro}

The rise of deep learning models (DNNs) has produced better prediction models for a plethora of fields, particularly for those that rely on unstructured data, such as computer vision and natural language processing (NLP)~\cite{peters2018deep, devlin2018bert}. In recent years, variants of these models have disseminated into many industrial applications, varying from image recognition to machine translation~\cite{szegedy2016rethinking, wu2016google, aharoni2019massively}. In NLP, they were also shown to produce better language models, and are being widely used both for language representation and for classification in nearly every sub-field~\cite{tshitoyan2019unsupervised, gao2019neural, lee2020biobert, feder2020active}.

While DNNs are very successful, this success has come at the expense of model explainability and interpretability. Understanding predictions made by these models is difficult, as their layered structure coupled with non-linear activations do not allow us to reason about the effect of each input feature on the model's output. In the case of text-based models this problem is amplified. Basic textual features are usually comprised of n-grams of adjacent words, but these features alone are limited in their ability to encode meaningful information conveyed in the text. While abstract linguistic concepts, such as topic or sentiment, do express meaningful information, they are usually not explicitly encoded in the model's input. \footnote{By concept, we refer to a higher level, often aggregated unit, compared to lower level, atomic input features such as words. Some examples of linguistic concepts are sentiment, linguistic register, formality or topics discussed in the text. For a more detailed discussion of concepts, see \cite{kim2018interpretability,goyal2019explaining}.} Such concepts might push the model towards making specific predictions, without being directly modeled and therefore interpreted. 

Effective concept-based explanations are crucial for the dissemination of DNN-based NLP prediction models in many domains, particularly in scientific applications to fields such as healthcare and the social sciences that often rely on model interpretability for deployment. Failing to account for the actual effect of concepts on text classifiers can potentially lead to biased, unfair, misinterpreted and incorrect predictions. As models are dependent on the data they are trained on, a bias existing in the data could potentially result in a model that under-performs when this bias no longer holds in the test set.

Recently, there have been many attempts to build tools that allow for DNN explanations and interpretations~\cite{ribeiro2016should, lundberg2017unified}, which have developed into a sub-field often referred to as Blackbox-NLP~\cite{ws2019acl}. These tools can be roughly divided into \textit{local explanations}, where the effect of a feature on the classifier's prediction for a specific example is tested, and \textit{global explanations}, which measure the general effect of a given feature on a classifier. A prominent research direction in DNN explainability involves utilizing network artifacts such as attention mechanisms, which are argued to provide a powerful representation tool~\cite{vaswani2017attention} to explain how certain decisions are made (but see \citep{jain2019attention} and \citep{wiegreffe2019attention} for a discussion of the actual explanation power of this approach). Alternatively, there have been attempts to estimate simpler, more easily-interpretable models, around test examples or their hidden representations~\cite{ribeiro2016should, kim2018interpretability}.

Unfortunately, existing model explanation tools often rely on local perturbations of the input and compute shallow correlations, which can result in misleading, and sometimes wrong, interpretations. This problem arises, for example, in cases where two concepts that can potentially explain the predictions of the model are strongly correlated with each other. An explanation model that only considers correlations cannot indicate if the first concept, the second concept or both concepts are in fact the cause of the prediction.

In order to illustrate the importance of causal and concept-based explanations, consider the example presented in Figure \ref{fig:example}, which will be our running example throughout the paper. Suppose we have a binary classifier, trained to predict the sentiment conveyed in news articles. Say we hypothesize that the choice of adjectives is driving the classification decision, something that has been discussed previously in computational linguistics~\cite{pang2002thumbs}. However, if the text is written about a controversial figure, it could be that the presence of its name, or the topics that it induces are what is driving the classification decision, and not the use of adjectives. The text in the figure is an example of such a case, where both adjectives and the mentioning of politicians seem to affect one another, and could be driving the classifier's prediction. Estimating the effect of Donald Trump's presence in the text on the predictions of the model is also hard, as this presence clearly affects the choice of adjectives, the other political figures mentioned in the text and probably many additional textual choices.

Notice, that an explanation model that only considers correlations might show that the mention of a political figure is strongly correlated with the prediction, leading to worries about the classifier having political bias. However, such a model cannot indicate whether the political figure is in fact the cause of the prediction, or whether it is actually the type of adjectives used that is the true cause of the classifier output, suggesting that the classifier is not politically biased. This highlights the importance of causal concept-based explanations.

A natural causal explanation methodology would be to generate counterfactual examples and compare the model prediction for each example with its prediction for the counterfactual. That is, one needs a controlled setting where it is possible to compute the difference between an actual observed text, and what the text would have been had a specific concept (e.g., a political figure) not existed in it.
Indeed, there have been some attempts to construct counterfactuals for generating local explanations. Specifically, \citet{goyal2019counterfactual} proposed changing the pixels of an image to those of another image classified differently by the classifier, in order to compute the effect of those pixels. However, as this method takes advantage of the spatial structure of images, it is hard to replicate their process with texts. \citet{vig2020causal} offered to use mediation analysis to study which parts of the DNN are pushing towards specific decisions by querying the language model. While their work further highlights the usefulness of counterfactual examples for answering causal questions in model interpretation, they create counterfactual examples manually, by changing specific tokens in the original example. Unfortunately, this does not support automatic estimation of the causal effect that high-level concepts have on model performance.

Going back to our example (Figure \ref{fig:example}), training a generative model to condition on a concept, such as the choice of adjectives, and produce counterfactual examples that only differ by this concept, is still intractable in most cases involving natural language (see Section \ref{subsec:generation} for a more detailed discussion). While there are instances where this seems to be improving~\cite{semeniuta2017hybrid, fedus2018maskgan}, generating a version of the example where a different political figure is being discussed while keeping other concepts unaffected is very hard~\cite{radford2018improving, radford2019language}. Alternatively, our key technical observation is that instead of generating a counterfactual text we can more easily generate a counterfactual textual representation, based on adversarial training.

It is important to note that it is not even necessarily clear what are the concepts that should be considered as the "generating concepts" of the text.\footnote{Our example also sheds more light on the nature of a concept. For example, if we train a figure classifier on the text after deleting the name of the political figure, it will probably still be able to classify the text correctly according to the figure it discusses. Hence, a concept is a more abstract entity, referring to an entire "semantic space/neighbourhood".} In our example we only consider adjectives and the political figure, but there are other concepts that generate the text, such as the topics being discussed, the sentiment being conveyed and others. The number of concepts that would be needed and their coverage of the generated text are also issues that we touch on below. The choice of such \textit{control concepts} depends on our model of the world, as in the \textit{causal graph} example presented in Figure \ref{fig:cg-example} (Section \ref{sec:causal}). In our experiments we control for such concepts, as our model of the world dictates both \textit{treated concepts} and \textit{control concepts}.\footnote{While failing to estimate the causal effect of a concept on a sentiment classifier is harmful, it pales in comparison to the potential harm of wrongfully interpreting clinical prediction models. In Appendix \ref{app:clinical} we give an example from the medical domain, where the importance of causal explanations has already been established \cite{zech2018variable}.}

In order to implement the above principles, in this paper we propose a model explanation methodology that manipulates the representation of the text rather than the text itself. By creating a text encoder that is not affected by a specific concept of interest, we can compute the \textit{counterfactual representation}. Our explanation method, which we name  \textit{Causal Model Explanation through Counterfactual Language Models} (\textit{CausaLM}), receives the classifier's training data and a concept of interest as input, and outputs the causal effect of the concept on the classifier in the test set. It does that by pre-training an additional instance of the language representation model employed by the classifier, with an adversarial component designed to "forget" the concept of choice, while keeping the other "important" (control) concepts represented. Following the additional training step, the representation produced by this counterfactual model can be used to measure the concept's effect on the classifier's prediction for each test example, by comparing the classifier performance with the two representations.

We start by diving into the link between causality and interpretability (Section \ref{sec:prev}). We then discuss how to estimate causal effects from observational data using language representations (Section \ref{sec:causal}): Defining the causal estimator (Section \ref{subsec:ci-lr} and \ref{subsec-treate}), discussing the challenges of producing counterfactual examples (Section \ref{subsec:generation}), and, with those options laid out, moving to describe how we can approximate counterfactual examples through manipulation of the language representation (Section \ref{subsec:lrm}). Importantly, our concept-based causal effect estimator does not require counterfactual examples -- it works solely with observational data.

\begin{figure}[b!]
    \centering
    \framebox{%
    \begin{varwidth}{0.90\textwidth}
    President \textcolor{red}{\textbf{Trump}} did his best imitation of \textcolor{red}{\textbf{Ronald Reagan}} at the State of the Union address, falling just short of declaring it Morning in America, the \textcolor{green}{\textbf{iconic}} imagery and message of a campaign ad that \textcolor{red}{\textbf{Reagan}} rode to re-election in 1984. \textcolor{red}{\textbf{Trump}} talked of Americans as pioneers and explorers; he lavished praise on members of the military, several of whom he recognized from the podium; he \textcolor{green}{\textbf{optimistically}} declared that the best is yet to come. It was a \textcolor{green}{\textbf{masterful}} performance -- but behind the \textcolor{green}{\textbf{sunny}} smile was the same old \textcolor{red}{\textbf{Trump}}: \textcolor{green}{\textbf{petty}}, \textcolor{green}{\textbf{angry}}, \textcolor{green}{\textbf{vindictive}} and \textcolor{green}{\textbf{deceptive}}. He refused to shake the hand of House Speaker \textcolor{blue}{\textbf{Nancy Pelosi}}, a snub she returned in kind by ostentatiously ripping up her copy of the President's speech at the conclusion of the address, in full view of the cameras.
    \end{varwidth}%
    }
    \caption{An example of a political commentary piece published at \url{https://edition.cnn.com}. Highlighted in \textcolor{blue}{\textbf{blue}} and \textcolor{red}{\textbf{red}} are names of political figures from the US Democratic and Republican parties, respectively. Adjectives are highlighted in \textcolor{green}{\textbf{green}}.}
    \label{fig:example}    
\end{figure}

To test our method, we introduce in Section \ref{sec:data} four novel datasets, three of which include counterfactual examples for a given concept. Building on those datasets, we present in Section \ref{sec:tasks} four cases where a BERT-based representation model can be modified to ignore concepts such as \textit{Adjectives}, \textit{Topics}, \textit{Gender} and \textit{Race}, in various settings involving sentiment and mood state classification (Section \ref{sec:tasks}). To prevent a loss of information on correlated concepts, we further modify the representation to remember such concepts while forgetting the concept whose causal effect is estimated. While in most of our experiments we test our methods in controlled settings, where the true causal concept effect can be measured, our approach can be used in the real-world, where such ground truth does not exist. Indeed, in our analysis we provide researchers with tools to estimate the quality of the causal estimator without access to gold standard causal information.

Using our newly created datasets, we estimate the causal effect of concepts on a BERT-based classifier utilizing our intervention method and compare to the ground truth causal effect, computed with manually created counterfactual examples (Section \ref{sec:results}). To equip researchers with tools for using our framework in the real-world, we provide an analysis of what happens to the language representation following the intervention, and discuss how to choose adversarial training tasks effectively (Section \ref{subsec:results-analysis}). As our approach relies only on interventions done prior to the supervised task training stage, it is not dependent on BERT's specific implementation and can be applied whenever a pre-trained language representation model is used. We also show that our counterfactual models can be used to mitigate unwanted bias in cases where its effect on the classifier can negatively affect outcomes.  Finally, we discuss the strengths and limitations of our approach, and propose future research directions at the intersection of causal inference and NLP model interpretation (Section \ref{sec:discussion}).

We hope that this research will spur more interest in the usefulness of causal inference for DNN interpretation and for creating more robust models, within the NLP community and beyond.

%% file: chapters/prev.tex
\section{Previous Work}
\label{sec:prev}

Previous work on the intersection of DNN interpretations and causal inference, specifically in relation to NLP is rare. While there is a vast and rich literature on each of those topics alone, the gap between interpretability, causality and NLP is only now starting to close~\cite{vig2020causal}. To ground our work in those pillars, we survey here previous work in each. Specifically, we discuss how to use causal inference in NLP~\cite{keith2020text}, and describe the current state of research on model interpretations and debiasing in NLP. Finally, we discuss our contribution in light of the relevant work.

\subsection{Causal Inference and NLP}
There is a rich body of work on causality and on causal inference, as it has been at the core of scientific reasoning since the writings of Plato and Aristotle~\cite{woodward2005making}. The questions that drive most researchers interested in understanding human behavior are causal in nature, not associational~\cite{pearl2009causal}. They require some knowledge or explicit assumptions regarding the data-generating process, such as the world model we describe in the causal graph presented in Figure \ref{fig:cg-example}. Generally speaking, causal questions cannot be answered using the data alone, or through the distributions that generate it~\cite{pearl2009causal}. 

Even though causal inference is widely used in the life and social sciences, it has not had the same impact on machine learning and NLP in particular~\cite{angrist2008mostly, dorie2019automated, gentzel2019case}. This can mostly be attributed to the fact that using existing frameworks from causal inference in NLP is challenging~\cite{keith2020text}. The high-dimensional nature of language does not easily fit into the current methods, specifically as the treatment whose effect is being tested is often binary~\cite{d2017overlap, athey2017estimating}. Recently, this seems to be changing, with substantial work being done on the intersection of causal inference and NLP~\cite{tan2014effect, fong2016discovery, egami2018make, wood2018challenges, veitch2019using}.

Specifically, researchers have been looking into methods of measuring other confounders via text~\cite{pennebaker2001linguistic, saha2019social}, or using text as confounders~\cite{johansson2016learning, de2016discovering, roberts2018adjusting}. In this strand of work, a confounder is being retrieved from the text and used to answer a causal question, or the text itself is used as a potential confounder, with its dimensionality reduced. Another promising direction is causally-driven representation learning, where the representation of the text is designed specifically for the purposes of causal inference. This is usually done when the treatment affects the text, and the model architecture is manipulated to incorporate the treatment assignment~\cite{roberts2014structural, roberts2018adjusting}. Recently,~\citet{veitch2019using} added to BERT's fine-tuning stage an objective that estimates propensity scores and conditional outcomes for the treatment and control variables, and used a model to estimate the treatment effect. As opposed to our work, they are interested in creating low-dimensional text embeddings that can be used as variables for answering causal questions, not in interpreting what affects an existing model.

While previous work from the causal inference literature used text to answer causal questions, to the best of our knowledge we are the first (except for~\cite{vig2020causal}) that are using this framework for causal model explanation. Specifically, we build in this research on a specific subset of causal inference literature, counterfactual analysis~\cite{pearl2009causality}, asking causal questions aimed at inferring what would have been the predictions of a given neural model had conditions been different. We present this counterfactual analysis as a method for interpreting DNN-based models, to understand what affects their decisions. By intervening on the textual representation, we provide a framework for answering causal questions regarding the effect of low and high level concepts on text classifiers without having to generate counterfactual examples.

~\citet{vig2020causal} also suggest using ideas from causality for DNN explanations, but focus on understanding how information flows through different model components, while we are interested in understanding the effect of textual concepts on classification decisions. They are dependant on manually constructed queries, such as comparing the language model's probability for a male pronoun to that of a female, for a given masked word. As their method can only be performed by manually creating counterfactual examples such as this query, it is exposed to all the problems involving counterfactual text generation (see Section \ref{subsec:generation}). Also, they do not compare model predictions on examples and their counterfactuals, and only measure the difference between the two queries, neither of which are the original text. In contrast, we propose a generalized method for providing a causal explanation for any textual concept, and present datasets where any causal estimator can be tested and compared to a ground truth. We also generate a language representation which approximates counterfactuals for a given concept of interest on each example, thus allowing for a causal model explanation without having to manually create examples.

\subsection{Model Interpretations and Debiasing in NLP}
Model interpretability is the degree to which a human can consistently predict the model's outcome~\cite{kim2016examples, doshi2017towards, lipton2018mythos}. The more easily interpretable a machine learning model is, the easier it is for someone to comprehend why certain decisions or predictions have been made. An explanation usually relates the feature values of an instance to its model prediction in a humanly understandable way, usually referred to as a \textit{local explanation}. Alternatively, it can be comprised of an estimation of the global effect of a certain feature on the model's predictions.

There is an abundance of recent work on model explanations and interpretations, especially following the rise of DNNs in the past few years~\cite{lundberg2017unified, ribeiro2016should}.~\citet{vig2020causal} divide interpretations in NLP into structural and behavioral methods. Structural methods try to identify the information encoded in the model's internal structure by using its representations to classify textual properties~\cite{adi2017fine, hupkes2018visualisation, conneau2018you}. For example,~\citet{adi2017fine} find that representations based on averaged word vectors encode information regarding sentence length. Behavioral methods evaluate models on specific examples that reflect an hypothesis regarding linguistic phenomena they capture~\cite{sennrich2017grammatical, isabelle2017challenge, naik2019exploring}.~\citet{sennrich2017grammatical}, for example, discover that neural machine translation systems perform transliteration better than models with byte-pair encoding (BPE) segmentation, but are worse in terms of capturing morphosyntactic agreement.

Both structural and behavioral methods generally do not offer ways to directly measure
the effect of the structure of the text or the linguistic concepts it manifests on model outcomes. They often rely on token level analysis, and do not account for counterfactuals. Still, there has been very little research in NLP on incorporating tools from causal analysis into model explanations~\cite{vig2020causal} (see above), something which lies at the heart of our work. Moreover, there's been, to the best of our knowledge, no work on measuring the effect of concepts on models' predictions in NLP (see~\citet{kim2018interpretability} and~\citet{goyal2019explaining} for a discussion in the context of computer vision).

Closely related to model interpretability, debiasing is a rising sub-field that deals with creating models and language representations that are unaffected by unwanted biases that might exist in the data~\cite{kiritchenko2018examining, elazar2018adversarial, gonen2019lipstick, ravfogel2020null}. DNNs are as good as the training data they are fed, and can often learn associations that are in direct proportion to the distribution observed during training~\cite{caliskan2017semantics}. While debiasing is still an ongoing effort, there are methods for removing some of the bias encoded in models and language representations~\cite{gonen2019lipstick}. Model debiasing is done through manipulation of the training data~\cite{kaushik2019learning}, by altering the training process~\cite{huang2019reducing} or by changing the model~\cite{gehrmann2019visual}. 

Recently,~\citet{ravfogel2020null} offered a method for removing bias from neural representations, by iteratively training linear classifiers and projecting the representations on their null-spaces. Their method does not provide causal model explanation, but instead reveals correlations between certain textual features and the predictions of the model. Particularly, it does not account for control concepts as we do, which makes it prone to overestimating the causal effect of the treatment concept (see Section \ref{sec:results} where we empirically demonstrate this phenomenon).

Our work is the first to provide datasets where bias can be computed directly by comparing predictions on examples and their counterfactuals. Comparatively, existing measures model bias using observational, rather than interventional measures~\cite{rudinger2017social, de2019bias, davidson2019racial, swinger2019biases, ravfogel2020null}. To compare methods for causal model explanations, the research community would require datasets, like those presented here, where we can intervene on specific textual features and test whether candidate methods can estimate their effect. In future we plan to develop richer, more complex datasets that would allow for even more realistic counterfactual comparisons.

%% file: chapters/causal.tex
\section{Causal Model Explanation}
\label{sec:causal}

While usually in scientific endeavors causal inference is the main focus, we rely here on a different aspect of causality - causal model explanation. That is, we attempt to estimate the causal effect of a given variable (also known as the \textit{treatment}) on the model's predictions, and present such effects to explain the observed behavior of the model. Here we formalize model explanation as a causal inference problem, and propose a method to do that through language representations.

We start by providing a short introduction to causal inference  and its basic terminology, focusing on its application to NLP. To ground our discussion within NLP, we follow the \textit{Adjectives} example from Section \ref{sec:intro} and present in Figure \ref{fig:cg-example} a \textit{casual diagram}, a graph that could describe the data-generating process of that example. Building on this graph, we discuss its connection to Pearl's \textit{structural causal model} and the \emph{do}-operator~\cite{pearl2009causal}. Typically, causal models are built for understanding real-world outcomes, while model interpretability efforts deal with the case where the classification decision is the outcome, and the intervention is on a feature present in the model's input. As we are the first, to the best of our knowledge, to propose a comprehensive causal framework for model interpretations in NLP, we link between the existing literature in both fields.

\subsection{Causal Inference and Language Representations}
\label{subsec:ci-lr}
\paragraph {Confounding Factors and the do-operator}
Continuing with the example from Section \ref{sec:intro} (presented in Figure \ref{fig:example}), imagine we observe a text $X$ and have trained a model to classify each example as either positive or negative, corresponding to the conveyed sentiment. We also have information regarding the \textit{Political Figure} discussed in the text, and tags for the parts of speech in it. Given a set of concepts, which we hypothesize might affect the model's classification decision, we denote the set of binary variables $C = \{C_j \in \{0,1\} | j \in \{0,1, \ldots, k \} \}$, where each variable corresponds to the existence of a predefined concept in the text, i.e., if $C_j=1$ then the $j$-th concept appears in the text. We further assume a pre-trained language representation model $\phi$ (such as BERT), and wish to assert how our trained classifier $f$ is affected by the concepts in $C$, where $f$ is a classifier that takes $\phi(X)$ as input and outputs a class $l \in L$. As we are interested in the effect on the probability assigned to each class by the classifier $f$, we measure the class probability of our output for an example $X$, and denote it for a class $l \in L$ as $z_{l}$. When computing differences on all $L$ classes, we use $\vec{z}(f(\phi(X)))$, the vector of all $z_{l}$ probabilities.

\begin{figure}[!htbp]
 \centering
 \includegraphics[width=0.5\linewidth]{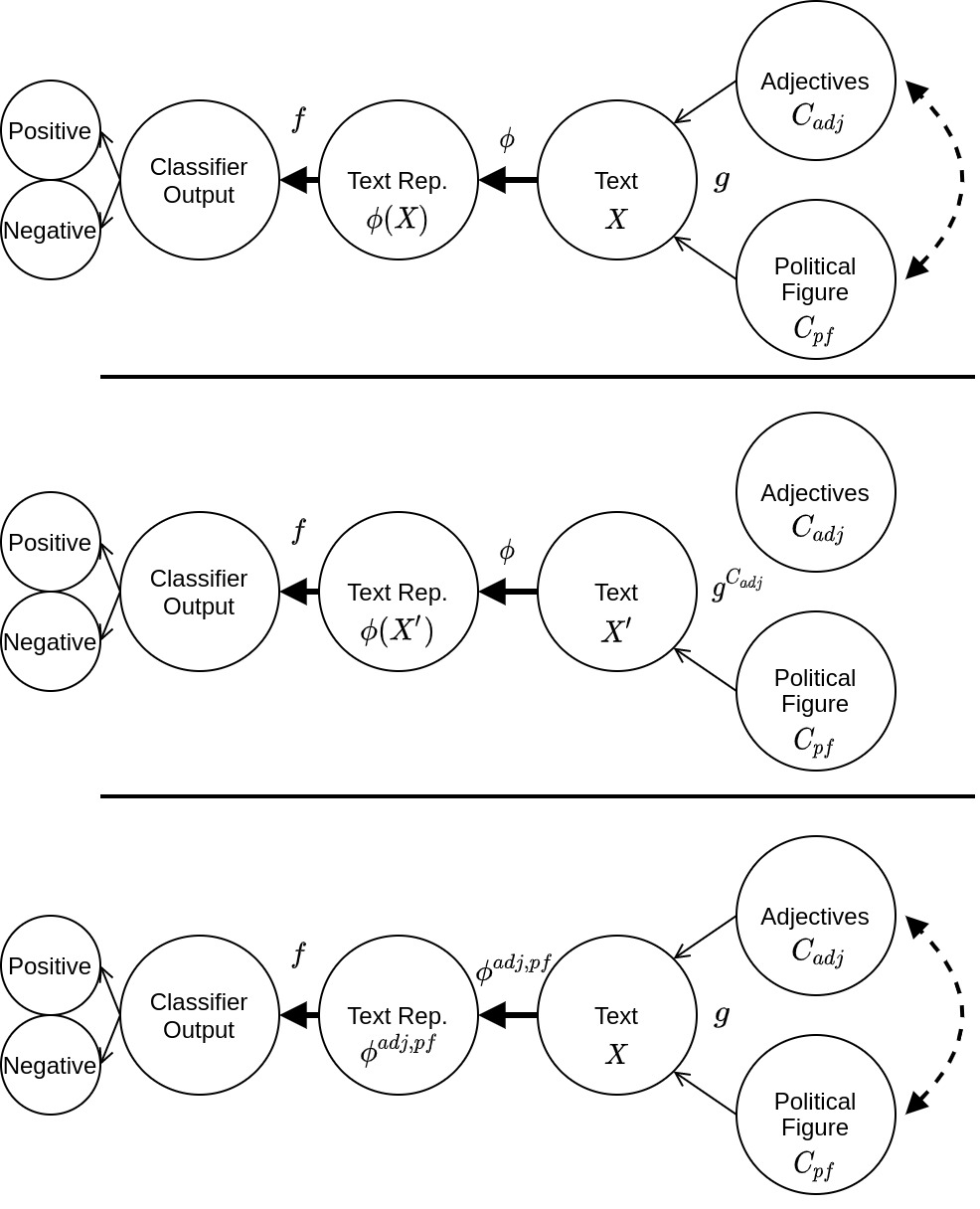}    
    \caption{Three causal graphs relating the concepts of \textit{Adjectives} and \textit{Political Figure}, texts, their representations and classifier output. The top graph describes the original data-generating process $g$. The middle graph describes the case of directly manipulating the text. In this case, using the generative process $g^{C_{adj}}$ allows us to generate a text $X'$ that is the same as $X$ but does not contain \textit{Adjectives}. The bottom graph describes our approach, where we manipulate the representation mechanism and not the actual text. The dashed edge indicates a possible hidden confounder of the two concepts.}
    \label{fig:cg-example}    
\end{figure}

Computing the effect of a concept $C_j$ on $\vec{z}(f(\phi(X)))$ seems like an easy problem. We can simply feed to our model examples with and without the chosen concepts, and compute the difference between the average $\vec{z}(\cdot)$ in both cases. For example, if our concept of interest is positive \textit{Adjectives}, we can feed the model with examples that include positive \textit{Adjectives} and examples that do not. Then, we can compare the difference between the averaged $\vec{z}(\cdot)$ in both sets and conclude that this difference is the effect of positive  \textit{Adjectives}. 

Now, imagine the case where the use of positive and negative \textit{Adjectives} is associated with the \textit{Political Figure} that is being discussed in the texts given to the model. An obvious example is a case where a political commentator with liberal-leaning opinions is writing about a conservative politician, or vice-versa. In that case, it would be reasonable to assume that the \textit{Political Figure} being discussed would affect the text through other concepts besides its identity. The author can then choose to express her opinion through \textit{Adjectives} or in other ways, and these might be correlated. In such cases, comparing examples with and without positive \textit{Adjectives} would result in an inaccurate measurement of their effect on the classification decisions of the model.\footnote{In fact, removing adjectives does provide a literal measurement of their impact, but it does not provide a measurement of the more abstract notion we are interested in (which is only partially expressed through the adjectives). Below we consider a baseline that does exactly this and demonstrate its shortcomings.}

The problem with our correlated concepts is that of \textit{confounding}. It is illustrated in the top graph of Figure \ref{fig:cg-example} using the example of \textit{Political Figure} and \textit{Adjectives}. In causal inference, a \textit{confounder} is a variable that affects other variables and the predicted label. In our case, the \textit{Political Figure} ($C_{pf}$) being discussed in the texts is a confounder of the \textit{Adjectives} concept, as it directly affects both $C_{adj}$ and $X$. As can be seen in this figure, we can think of texts as originating from a list of concepts. While we plot only two, \textit{Adjectives} and \textit{Political Figure}, there could be many concepts generating a text. We denote the potential confoundedness of the concepts by dashed arrows, to represent that one could affect the other or that they have a common cause.

Alternatively, if it was the case that a change of the \textit{Political Figure} would not affect the usage of \textit{Adjectives} in the text, we could have said that $C_{adj}$ and $C_{pf}$ are not confounded. This is the case where we could intervene on $C_{adj}$, such as by having the author write a text without using positive \textit{Adjectives}, without inducing a text that contains a different \textit{Political Figure}. In causal terms, this is the case where:
\begin{equation}
\vec{z}(f(\phi(X)| \emph{do}(C_{adj}))) = \vec{z}(f(\phi(X)| C_{adj}))
\end{equation}

Where $\emph{do}(C_{adj})$ stands for an external intervention that compels the change of $C_{adj}$. In contrast, the class probability distribution $\vec{z}(f(\phi(X)| C_{adj}))$ represents the distribution resulting from a passive observation of $C_{adj}$, and rarely coincides with $\vec{z}(f(\phi(X)| \emph{do}(C_{adj})))$. Indeed, the passive observation setup relates to the probability that the sentiment is positive given that positive adjectives are used. In contrast, the external intervention setup relates to the probability that the sentiment is positive after all the information about positive adjectives has been removed from a text that originally (pre-intervention) conveyed positive sentiment.

\paragraph {Counterfactual Text Representations}
The act of manipulating the text to change the \textit{Political Figure} being discussed or the \textit{Adjectives} used in the text is derived from the notion of \textit{counterfactuals}. In the \textit{Adjectives} example (presented in Figure \ref{fig:example}), a counterfactual text is such an instance where we intervene on one concept only, holding everything else equal. It is the equivalent of imagining what could have been the text, had it been written about a different \textit{Political Figure}, or about the same \textit{Political Figure} but with different \textit{Adjectives}.

In the case of \textit{Adjectives}, we can simply detect all of them in the text and change them to a random alternative, or delete them altogether.\footnote{This would still require the modeler to control some confounding concepts, as \textit{Adjectives} could be correlated with other variables (such as some \textit{Adjectives} used to describe a specific politician).} For the concept highlighting the \textit{Political Figure} being discussed this is much harder to do manually, as the chosen figure induces the topics being described in the text and is hence likely to affect other important concepts that generate the text.

Intervening on \textit{Adjectives} as presented in the middle graph of Figure \ref{fig:cg-example} relies on our ability to create a conditional generative model, one that makes sure a certain concept is or is not represented in the text. Since this is often hard to do (see Section \ref{subsec:generation}), we propose a solution that is based on the language representation $\phi(X)$. As shown in the bottom causal graph of Figure \ref{fig:cg-example}, we assume that the concepts generate the representation $\phi(X)$ directly. This approximation shares some similarities with the idea of \textit{Process Control} described in~\citet{pearl2009causal}. While Pearl presents \textit{Process Control} as the case of intervening on the process affected by the treatment, it is not discussed in relation to language representations or model interpretations. Interventions on the process that is generating the outcomes are also discussed in Chapter 4 of~\citet{bottou2013counterfactual}, in the context of multi-armed bandits and reinforcement learning.

By intervening on the language representation, we attempt to bypass the process of generating a text given that a certain concept should or should not be represented in that text. We take advantage of the fact that modern NLP systems use pre-training to produce a language representation, and generate a counterfactual language representation $\phi^{C}(X)$ that is unaffected by the existence of a chosen concept $C$. That is, we try to change the language representation such that we get for a binary $C$: 
\begin{equation}
\vec{z}(f(\phi^{C}(X))) = \vec{z}(f(\phi^{C}(X')))
\end{equation}
Where $X$ and $X'$ are identical for every generating concept, except for the concept $C$, on which they might or might not differ. In Section \ref{sec:meth}, we discuss how we intervene in the fine-tuning stage of the language representation model (BERT in our case) to produce the counterfactual representation using an adversarial component. 

We now formally define our causal effect estimator. We start with the definition of the standard \textit{Average Treatment Effect} (ATE) estimator from the causal literature. We next formally define the \textit{causal concept effect} (CaCE), first introduced in~\citet{goyal2019explaining} in the context of computer vision. We then define the Example-based Average Treatment Effect (EATE), a related causal estimator for the effect of the existence of a concept on the classifier. The process required to calculate EATE is presented in the middle graph of Figure~\ref{fig:cg-example}, and requires a conditional generative model. In order to avoid the need in such a conditional generative model, we follow the bottom graph of Figure~\ref{fig:cg-example} and use an adversarial method, inspired by the idea of \textit{Process Control} that was first introduced by~\citet{pearl2009causality}, to intervene on the text representation. We finally define the \textit{Textual Representation-based Average Treatment Effect} (TReATE), which is estimated using our method, and compare it to the standard ATE estimator.\footnote{In appendix \ref{app:alt-graph} we discuss alternative causal graphs that describe different types of relationships between the involved variables. We also discuss the estimation of causal effects in such cases and briefly touch on the selection of the appropriate causal graph for a given problem.}

\subsection{The Textual Representation-based Average Treatment Effect (TReATE)}
\label{subsec-treate}

When estimating causal effects, researchers commonly measure the \textit{average treatment effect}, which is the difference in mean outcomes between the treatment and control groups. Using \emph{do}-calculus~\cite{pearl1995causal}, we can define it in the following way:
\begin{definition}[Average Treatment Effect (ATE)] 
The average treatment effect of a binary treatment $T$ on the outcome $Y$ is:
\begin{equation}\label{eq:ATE}
    \text{ATE}_{T} = \mathbb{E}\left[Y|do(T=1)\right] - \mathbb{E}\left[Y|do(T=0)\right]
\end{equation}
\end{definition}

Following the notations presented in the beginning of Section \ref{subsec:ci-lr}, we define the following Structural Causal Model (SCM, \citet{pearl2009causality}) for a document $X$:

\begin{align}\label{eq:scm1}
    & (C_0, C_1, \ldots, C_k) =  h(\epsilon_C) \nonumber \\
    & X = g(C_0, C_1, \ldots, C_k,\epsilon_X) \nonumber \\
    & C_{j} \in \{0, 1 \} \text{, } \forall j \in K \nonumber \\
\end{align}
Where, as is standard in SCMs, $\epsilon_C$ and $\epsilon_X$ are independent variables. The function $h$ is the generating process of the concept variables from the random variable $\epsilon_C$ and is not the focus here. The SCM in Equation \eqref{eq:scm1} makes an important assumption, namely that it is possible to intervene atomically on $C_j$, the \textit{treated concept} (TC), while leaving all other concepts untouched. 

We denote expectations under the interventional distribution by the standard \emph{do}-operator notation $\mathbb{E}_g \left[ \cdot | do(C_j=a)\right]$, where the subscript $g$ indicates that this expectation also depends on the generative process $g$. We can now use these expectations to define \textit{CaCE}: 

\begin{definition}[Causal Concept Effect (CaCE)~\cite{goyal2019explaining}] 
The causal effect of a concept $C_j$ on the class probability distribution $\vec{z}$ of the classifier $f$ trained over the representation $\phi$ under the generative process $g$ is:
\begin{equation}
    \text{CaCE}_{C_j} =  \langle \mathbb{E}_g\left[\vec{z}\big(f(\phi(X))\big) | do(C_j=1) \right] - \mathbb{E}_{g}\left[\vec{z}\big(f(\phi(X))\big) | do(C_j=0) \right] \rangle
\end{equation}
\end{definition}
Where $\langle \rangle$ is the $l_1$ norm: A summation over the absolute values of vector coordinates.\footnote{For example, for a three class prediction problem, where the model's probability class distribution for the original example is $(0.7, 0.2, 0.1)$, while for the counterfactual example it is $(0.5, 0.1, 0.4)$, \textit{${CaCE}_{C_j}$} is equal to: $|0.7-0.5|+|0.2-0.1|+|0.1-0.4| = 0.2+0.1+0.3=0.6$.}

\textit{CaCE} was designed to test how a model would perform if we intervene and change a value of a specific concept (e.g., if we changed the hair color of a person in a picture from blond to black). Here we address an alternative case, where some concept exists in the text and we aim to measure the causal effect of its existence on the classifier. As can be seen in the middle causal graph of Figure \ref{fig:cg-example}, this requires an alternative data-generating process $g^{C_0}$, which is not affected by the concept $C_0$. Using $g^{C_0}$, we can define another SCM that describes this relationship:

\begin{align}\label{eq:scm2}
    & (C_0, C_1, \ldots, C_k) =  h(\epsilon_C) \nonumber \\
    & X' = g^{C_0}(C_1, \ldots, C_k,\epsilon_X') \nonumber \\
    & C_{j} \in \{0, 1 \} \text{, } \forall j \in K \nonumber \\
\end{align}
Where $X'$ is a counterfactual example generated by $g^{C_0}(C_1=c_1,\ldots,C_k=c_k, \epsilon_X')$. With $g^{C_0}$, we want to generate texts that use $(C_1=c_1,\ldots,C_k=c_k)$ in the same way that $g$ does, but are as if $C_0$ never existed.  Using this SCM, we can compute the Example-based Average Treatment Effect (\textit{EATE}):

\begin{definition}[Example-based Average Treatment Effect (EATE)] 
The causal effect of a concept $C_j$ on the class probability distribution $\vec{z}$ of the classifier $f$ under the generative processes $g, g^{C_j}$ is:
\begin{equation}
    \text{EATE}_{C_j} =  \langle \mathbb{E}_{g^{C_j}}\left[\vec{z}\big(f(\phi(X'))\big) \right] - \mathbb{E}_{g}\left[\vec{z}\big(f(\phi(X))\big) \right] \rangle
\end{equation}
\end{definition}

Implementing \textit{EATE} requires counterfactual example generation, as shown in the middle graph of Figure \ref{fig:cg-example}. As this is often intractable in NLP (see Section \ref{subsec:generation}), we do not compute \textit{EATE} here. We instead generate a counterfactual language representation, a process which is inspired by the idea of \textit{Process Control} introduced by~\citet{pearl2009causality} for dynamic planning. This is the case where we can only control the process generating $\phi(X)$ and not $X$ itself. 

Concretely, using the middle causal graph in Figure \ref{fig:cg-example}, we could have generated two examples $X_1 = g^{C_0}(C_1=c_1,\ldots,C_k=c_k,\epsilon_{X'}=\epsilon_{x'})$ and $X_2 = g^{C_0}(C_1=c_1,\ldots,C_k=c_k,\epsilon_{X'}=\epsilon_{x'})$ where $C_0=1$ for $X_1$ and $C_0=0$ for $X_2$, and have that $X_1=X_2$ because the altered generative process $g^{C_0}$ is not sensitive to changes in $C_0$. Notice that we require that $g^{C_0}$ would be similar to $g$ in the way the concepts $(C_1,\ldots,C_k)$ generate the text, because otherwise any degenerate process will do. Alternatively, in the case where we do not have access to the desired conditional generative model, we would like for the two examples $\bar{X_1} = g(C_0=1,C_1=c_1,\ldots,C_k=c_k,\epsilon_X=\epsilon_x)$ and $\bar{X_2} = g(C_0=0,C_1=c_1,\ldots,C_k=c_k,\epsilon_X=\epsilon_x)$, to have that $\phi^{C_0}(\bar{X_1}) = \phi^{C_0}(\bar{X_2})$. That is, we follow the bottom graph from Figure \ref{fig:cg-example}, and intervene only on the language representation $\phi(X)$ such that the resulting representation, $\phi^{C_0}(X)$, is insensitive to $C_0$ and is similar to $\phi$ in the way the concepts $(C_1,\ldots,C_k)$ are represented. Following this intervention, we compute the \textit{Textual Representation-based Average Treatment Effect} (\textit{TReATE}).

\begin{definition}[Textual Representation-based Average Treatment Effect (TReATE)] 
The causal effect of a concept $C_j$, controlling for concept $C_{m}$, on the class probability distribution $\vec{z}$ of the classifier $f$ under the generative process $g$ is:
\begin{equation}\label{eq:TReATE_CC}
    \text{TReATE}_{C_j,C_m} = \langle \mathbb{E}_g\left[\vec{z}\big(f(\phi(X))\big)\right]  - \mathbb{E}_g\left[\vec{z}\big(f(\phi^{C_j,C_m}(X))\big)\right] \rangle 
\end{equation}
\end{definition}
Where $\{ C_j,C_m \}$ denotes the concept (or concepts) $C_j$ whose effect we are estimating, and $C_m$ the potentially confounding concept (or concepts) we are controlling for. In order to not overwhelm the notation, whenever we use only one concept in the superscript it is the concept whose effect is being estimated, and not the confounders.

In our framework, we would like to use the tools defined here to measure the casual effect of one or more concepts $\{ C_0, C_1, \cdots ,  C_k \}$ on the predictions of the classifier $f$. We will do that by measuring \textit{TReATE}, which is a special case of the \textit{average treatment effect} (ATE) defined in Equation \ref{eq:ATE}, where the intervention is performed via the textual representation. While \textit{ATE} is usually used to compute the effect of interventions in randomized experiments, here we use \textit{TReATE} to explain the predictions of a text classification model in terms of concepts.

\subsection{Representation-Based Counterfactual Generation}
\label{sec:meth}
    
We next discuss the reason we choose to intervene through the language representation mechanism, as an alternative to synthetic example generation. We present two existing approaches for generating such synthetic examples and explain why they are often implausible in NLP. We then introduce our approach, an intervention on the language representation, designed to ignore a particular set of concepts while preserving the information from another set of concepts. Finally, we describe how to perform this intervention using the counterfactual language representation.

\paragraph{Generating Synthetic Examples}
\label{subsec:generation}

Comparing model predictions on examples to the predictions on their counterfactuals is what allows the estimation of causal explanations. Without producing a version of the example that does not contain the treatment (i.e concept or feature of interest), it would be hard to ascertain whether the classifier is using the treatment or other correlated information~\cite{kaushik2019learning}. To the best of our knowledge, there are two existing methods for generating counterfactual examples: manual augmentation and automatic generation using generative models.

Manual augmentation can be straight-forward, as one needs to manually change every example of interest to reflect the absence or presence of a concept of choice. For example, when measuring the effect of \textit{Adjectives} on a sentiment classifier, a manual augmentation could include changing all positive \textit{Adjectives} into negative ones, or simply deleting all \textit{Adjectives}. While such manipulations can sometime be easily done with human annotators, they are costly and time consuming and therefore implausible for large datasets. Also, in cases such as the clinical note example presented in Figure \ref{fig:example-clinical}, it would be hard to manipulate the text such that it uses a different writing style, making it even harder to manually create the counterfactual text.

Using generative models has been recently discussed in the case of images~\cite{goyal2019explaining}. In this paper, Goyal et al. propose using a conditional generative model, such as a conditional VAE~\cite{lorberbom2019direct}, to create counterfactual examples. While in some cases, such as those presented in their paper, it might be plausible to generate counterfactual examples, in most cases in NLP it is still too hard to generate realistic texts with conditional generative models~\cite{lin2017adversarial, che2017maximum, rajeswar2017adversarial, guo2018long}. Also, for generating local explanations it is required to produce a counterfactual for each example such that all the information besides the concept of choice is preserved, something that is even harder than producing two synthetic examples, one from each concept class, and comparing them.

As an alternative to manipulating the actual text, we propose to intervene on the language representation. This does not require generating more examples, and therefore does not depend on the quality of the generation process. The fundamental premise of our method is that comparing the original representation of an example to this counterfactual representation is a good approximation of comparing an example to that of a synthetic counterfactual example that was properly manipulated to ignore the concept of interest.

\paragraph{Interventions on Language Representation Models}
\label{subsec:lrm}
Since the introduction of pre-trained word-embeddings, there has been an explosion of research on choosing pre-training tasks and understanding their effect~\cite{jernite2017discourse, logeswaran2018efficient, Ziser:18, dong2019unified, chang2019language, sun2019fine, rotman2019deep}. The goal of this process is to generate a representation that captures valuable information for solving downstream tasks, such as sentiment classification, entity recognition and parsing. Recently, there has also been a shift in focus towards pre-training contextual language representations~\cite{liu2019roberta, yang2019xlnet}.

Contextual embedding models typically follow three stages: \textbf{(1)} Pre-training: Where a DNN (encoder) is trained on a massive unlabeled dataset to solve self-supervised tasks; \textbf{(2)} Fine-tuning: An optional step, where the encoder is further trained on different tasks or data; and \textbf{(3)} Supervised task training: Where task specific layers are trained on labeled data for a downstream task of interest. 

\begin{figure}[htbp]
 \centering
 \includegraphics[width=0.8\linewidth]{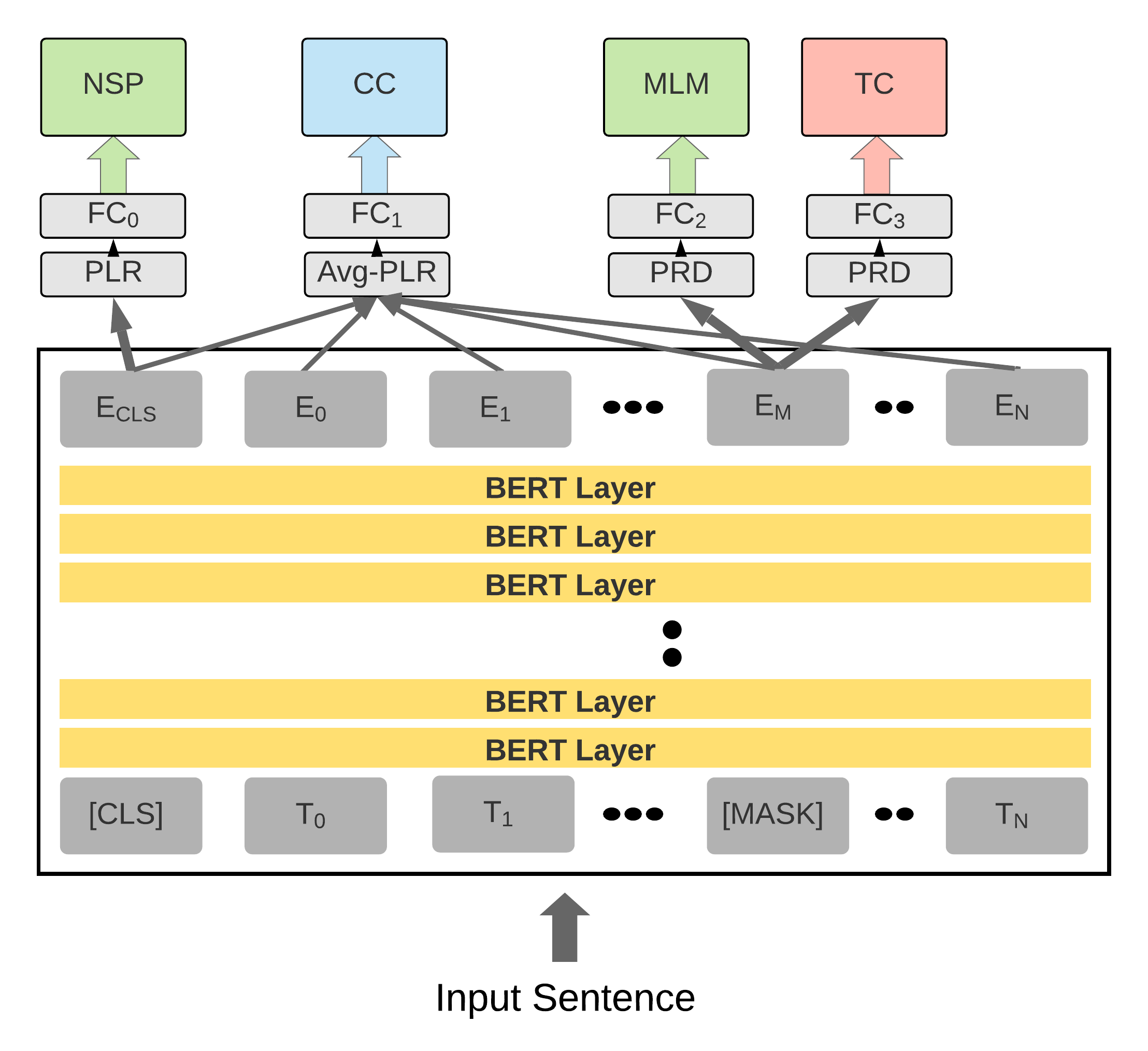}    
    \caption{An illustration of our Stage 2 fine-tuning procedure for our counterfactual representation model (\textit{BERT-CF}). In this representative case, we add a task, named \textit{Treated Concept} (TC), which is trained adversarially. This task is designed to "forget" the effect of the treated concept, as in the $IMA$ adversarial task discussed in Section \ref{sec:tasks}. To control for a potential confounding concept (i.e., to "remember" it), we add the \textit{Control Concept} (CC) task, which predicts the presence of this concept in the text, as in the $PF$ task discussed below. $PRD$ and $PLR$ stand for BERT's prediction head and the pooler head respectively, $AVG-PLR$ for an average pooler head, FC is a fully connected layer, and $[MASK]$ stands for masked tokens embeddings. $NSP$ and $MLM$ are BERT's next prediction and masked language model objectives. The results of this training stage is our counterfactual \textit{BERT-CF} model.}
    \label{fig:BERT-CF}    
\end{figure}

Our intervention is focused on Stage 2. In this stage, we continue training the encoder of the model on the tasks it was pre-trained on, but add auxiliary tasks, designed to forget some concepts and remember others.\footnote{Continued pre-training has shown useful in NLP more generally  \citep{gururangan2020don, gardner2020evaluating}.} In Figure \ref{fig:BERT-CF} we present an example of our proposed Stage 2, where we train our model to solve the original BERT's \textit{Masked Language Model} ($MLM$) and \textit{Next Sentence Prediction} ($NSP$) tasks, along with a \textit{Treated Concept} objective, denoted in the figure as \textit{TC}. In order to preserve the information regarding a potentially confounding concept, we use an additional task denoted in the figure as \textit{CC}, for \textit{Controlled Concept}.

To illustrate our intervention, we can revisit the \textit{Adjectives} example of Figure \ref{fig:example}, and consider a case where we want to test whether their existence in the text affects the classification decision. To be able to estimate this effect, we traditionally would have to produce for each example in the test-set an equivalent example that does not contain \textit{Adjectives}. In terms of our intervention on the language representation, we should be able to produce a representation that is unaffected by the existence of \textit{Adjectives}, meaning that the representation of a sentence that contains \textit{Adjectives} would be identical to that of the same sentence where \textit{Adjectives} are excluded. Taking that to the fine-tuning stage, we could use adversarial training to "forget" \textit{Adjectives}.

Concretely, we add to BERT's loss function a negative term for the target concept and a positive term for each control concept we consider. As shown in Equation \ref{eq:loss}, in the case of the example from Figure \ref{fig:example}, this would entail augmenting the loss function with two terms: adding the loss for the \textit{Political Figure} classification $PF$ (the \textit{CC} head), and subtracting that of the \textit{Is Masked Adjective} ($IMA$) task (the \textit{TC} head). As we are using the $IMA$ objective term in our \textit{Adjectives} experiments (Section \ref{sec:tasks}), and not only in the running example, we describe the task below. For the \textit{Political Figure} ($PF$) concept, we could simply use a classification task where for each example we predict the political orientation of the politician being discussed.\footnote{For the \textit{CC} objective, we can add any of the classification tasks suggested above for $PF$ (\textit{CC}), following the definition of the world model (i.e., the causal graph) the researcher is assuming.} With those tasks added to the loss function, we have that:

\begin{align}\label{eq:loss}
\mathcal{L} (\theta_{bert},\theta_{mlm},\theta_{nsp},\theta_{cc},\theta_{tc}) =
& \frac{1}{n} \Big(\sum_{i=1}^n \mathcal{L}_{mlm}^i(\theta_{bert},\theta_{mlm}) \nonumber 
& + \sum_{i=1}^n \mathcal{L}_{nsp}^i(\theta_{bert},\theta_{nsp}) \nonumber \\
& + \sum_{i=1}^n \mathcal{L}_{cc}^i(\theta_{bert},\theta_{cc})
& - \lambda \sum_{i=1}^{n} \mathcal{L}_{tc}^i(\theta_{bert},\theta_{tc}) \Big) 
\end{align}
Where $\theta_{bert}$ denotes all of BERT's parameters, except those devoted to $\theta_{mlm},\theta_{nsp},\theta_{tc}$ and $\theta_{cc}$. $\lambda$ is a hyper-parameter which controls the relative weight of the adversarial task. 
One way of implementing the $IMA$ \textit{TC} head is inspired by BERT's $MLM$ head. That is, masking \textit{Adjectives} and \textit{Non-adjectives}, then predicting whether the masked token is an adjective. Following the \textit{gradient reversal} method (\citet{ganin2016domain}, henceforth DANN),\footnote{See equations $9-10$ and $13-15$ in~\citet{ganin2016domain}.} we add this task with a layer which leaves the input unchanged during forward propagation, yet reverses its corresponding gradients by multiplying them with a negative scalar ($-\lambda$) during back propagation. 

The core idea of DANN is to reduce the domain gap, by learning common representations that are indistinguishable to a domain discriminator \cite{Ghosal2020KinGDOMKD}. In our model, we replace the domain discriminator with a discriminator that discriminates examples with the treated concept from examples that does not have that concept. Following DANN, we optimize the underlying BERT representations jointly with classifiers operating on these representations: The task classifiers perform the main task of the model ($\mathcal{L}_{mlm}$, $\mathcal{L}_{nsp}$ and $\mathcal{L}_{cc}$ in our objective) and the treatment concept classifier discriminates between those masked tokens which are adjectives and those which are not (the $\mathcal{L}_{tc}$ term in our objective). 
While the parameters of the classifiers ($\theta_{mlm}$, $\theta_{nsp}$, $\theta_{cc}$, $\theta_{tc}$) are optimized in order to minimize their training error,
the language encoder parameters ($\theta_{bert}$) are optimized in order to minimize the loss of the task classifiers ($\mathcal{L}_{mlm}$, $\mathcal{L}_{nsp}$ and $\mathcal{L}_{cc}$) and to maximize the loss of the treatment concept classifier ($\mathcal{L}_{tc}$). Concretely in our case, the parameters of the underlying language representation $\theta_{bert}$ are simultaneously optimized in order to minimize the $MLM$, $NSP$ and $PF$ loss functions and maximize the $IMA$ loss. \textit{Gradient reversal} hence encourages an adjective-invariant language representation to emerge. For more information about the adversarial multi-task min-max optimization dynamics, and the emergent concept-invariant language representations, see \citet{xie2017controllable}.

While the \textit{gradient reversal} method is widely implemented throughout the domain adaptation literature \cite{ramponi2020neural}, it has also been previously shown that it can be at odds with the model's main prediction objective \cite{elazar2018adversarial}. However, we implement it in our model's training process in a different way than in most previous literature. We use this method as part of the language model fine-tuning stage, which is independent of and precedes the downstream prediction task's objective training. Therefore, our adversarial task's objective is not directly at odds with the downstream model's prediction objective.

Having optimized the loss functions presented in Equation \ref{eq:loss}, we can now use the resulting counterfactual representation model and compute the \textit{individual treatment effect} (ITE) on an example as follows. We compute the predictions of two different models: One that employs the original BERT, that has not gone through our counterfactual fine-tuning, and one that employs the counterfactual BERT model (BERT-CF). \textit{The Textual Representation-based ITE} (TRITE) is then the average of the absolute differences between the probabilities assigned to the possible classes by these models.
As \textit{TReATE} is presented in Equation \ref{eq:TReATE_CC} in expectation form, we compute our estimated $\widehat{TReATE}$ by summing over $\widehat{TRITE}$ for the set of all test-set examples, $I$:

\begin{align} \label{eq:TReATE_hat}
    \widehat{TReATE}_{TC,CC} = &\frac{1}{|I|}\sum_{i \in I} \widehat{TRITE}_{TC,CC}^{i} = \nonumber \\
    & \frac{1}{|I|}\sum_{i \in I} \langle \vec{z}\big(f(\phi^{TC,CC}(X=x_i))\big) - \vec{z}\big( f(\phi(X=x_i))\big) \rangle
\end{align}
Where $x_i$ is the specific example, $\phi$ is the original language representation model and $\phi^{TC,CC}$ is the counterfactual \textit{BERT-CF} representation model, where the intervention is such that \textit{TC} has no effect and \textit{CC} is preserved. $\vec{z}\big( f(\phi(X))\big)$ is the class probability distribution of the classifier $f$ when using $\phi$ as the representation model for example $X$.

%% file: chapters/data.tex
\section{Data}
\label{sec:data}

When evaluating a trained classification model, we usually have access to a test-set, consisting of manually labeled examples that the model was not trained on, and can hence be used for evaluation. Estimating causal effects is often harder in comparison, as we do not have access to the ground truth. In the case of causal inference, we can generally only identify effects if our assumptions on the data-generating process, such as those presented in Figure \ref{fig:cg-example}, hold. This means that at the core of our causal model explanation paradigm is the availability of a causal graph that encodes our assumptions about the world. Notice, however, that non-causal explanation methods that do not make assumptions about the world are prone to finding arbitrary correlations, a problem that we are aiming to avoid with our method.

To allow for ground-truth comparisons and to spur further research on causal inference in NLP, we propose here four cases where causal effects can be estimated. In three out of those cases, we have constructed datasets with counterfactual examples so that the causal estimators can be compared to the ground truth. We start here by introducing the datasets we created and discuss the choices made in order to allow for proper evaluation.  Section \ref{subsec:sentiment_reviews} describes the sentiment analysis data with the \textit{Adjectives} and \textit{Topics} concepts, while Section \ref{subsec:eeec}  describes the EEEC dataset for mood classification with the \textit{Gender} and \textit{Race} concepts. Section \ref{sec:tasks} presents the tasks for which we estimate the causal effect, and the resulting experiments.~\footnote{Our datasets are available at: \url{https://www.kaggle.com/amirfeder/causalm}.}

\subsection{Product and Movie Reviews}
\label{subsec:sentiment_reviews}
Following the running example of Section \ref{sec:intro}, we start by looking for prominent sentiment classification datasets. Specifically, we look for datasets where the domain entails a rich description where \textit{Adjectives} could play a vital role. With enough variation in the structure and length of examples, we hope that \textit{Adjectives} would have a significant effect. Another key aspect is the number of training examples. To be able to amplify the correlation between the treated concept (\textit{Adjectives}) and the label, we need to be able to omit some training examples. For instance, if we omit most of the positive texts describing a \textit{Political Figure}, we can create a correlation between the negative label and that politician. We need a dataset that will allow us to do that and still have enough training data to properly train modern DNN classification models.

We also wish to estimate the causal effect of the concept of \textit{Topics} on sentiment classification (see Section \ref{sec:tasks} for an explanation on how we compute the topic distribution). To be able to observe the causal effect of \textit{Topics}, some variation is required in the \textit{Topics} discussed in the texts. For that, we use data originating from several different domains, where different, unrelated products or movies are being discussed. In this section we focus on the description of the dataset we have generated, and explain how we manipulate the data in order to generate various degrees of concept-label correlations.

Considering these requirements and the concepts for which we wish to estimate the causal effect on model performance, we choose to combine two datasets, spanning five domains. The product dataset we choose is widely used in the NLP domain adaptation literature, and is taken from~\citet{blitzer2007biographies}. It contains four different domains: \textit{Books}, \textit{DVD}, \textit{Electronics} and \textit{Kitchen Appliances}. The movie dataset is the IMDB movie review dataset, taken from~\citet{maas2011learning}. In both datasets, each example consists of a review and a rating ($0$-$5$ stars). Reviews with $rating > 3$ were labeled positive, those with $rating < 3$ were labeled negative, and the rest were discarded because their polarity was ambiguous. The product dataset is comprised of $1,000$ positive and $1,000$ negative examples for each of the four domains, for a total of $4,000$ positive and $4,000$ negative reviews. The \textit{Movies} dataset is comprised of $25,000$ negative and $25,000$ positive reviews. To construct our combined dataset, we randomly sample $1,000$ positive and $1,000$ negative reviews from the \textit{Movies} dataset and add these alongside the product dataset reviews. Our final combined dataset amounts to a total of $10,000$ reviews, balanced across all five domains and both labels.

We tag all examples in both datasets for the Part-of-Speech (\textit{PoS}) of each word with the automatic tagger available through \textit{spaCy},\footnote{\url{https://spacy.io/}} and use the predicted labels as ground truth. For each example in the combined dataset, we generate a counterfactual example for \textit{Adjectives}. That is, for each example we create another instance where we delete all words that are tagged as \textit{Adjectives}, such that for the example: "\textit{It's a lovely table}", the counterfactual example will be: "\textit{It's a table}". Finally, we count the number of \textit{Adjectives} and other \textit{PoS} tags, and create a variable indicating the ratio of \textit{Adjectives} to \textit{Non-adjectives} in each example, which we use in Section \ref{sec:tasks} to bias the data.

For the \textit{Topic} concepts, we train an LDA topic model~\cite{blei2003latent}\footnote{Using the \textit{gensim} library~\cite{rehurek2010gensim}.} on all the data in our combined dataset and optimize the number of topics for maximal coherence~\cite{lau2014machine}, resulting in a set of $T=50$ topics. For each of the five domains we then search for the \textit{treatment concept} topic $t_{TC}$, which we define as the topic which is relatively most associated with that domain, i.e., the topic with the largest difference between the probability assigned to examples from that domain and the probability assigned to examples outside of that domain, using the following equation:
\begin{equation}\label{eq:topic}
    t_{TC}(d) = \argmax_{t \in T} \big( \frac{1}{|I_{d_+}|} \sum_{i \in I_{d_+}} \theta_{t}^{i} - \frac{1}{|I_{d_-}|} \sum_{i \in I_{d_-}} \theta_{t}^{i} \big)
\end{equation}
Where $d$ is the domain of choice, $t$ is a topic from the set of topics $T$, $\theta_{t}$ is the probability of topic $t$ and $I_{domain}$ is the set of examples for a given domain. $I_{d+}$ is the set of examples in domain $d$, and $I_{d-}$ the set of examples outside of domain $d$. After choosing $t_{TC}$, we exclude it from $T$ and use the same process to choose $t_{CC}$, our \textit{control concept} topic. 

For each \textit{Topic}, we also compute the median probability on all examples, and define a binary variable indicating for each example whether the \textit{Topic} probability is above or below its median. This binary variable can then be used for the \textit{TC} and \textit{CC} tasks described in Section \ref{sec:tasks}.

In Table \ref{tab:sentiment-data} we present some descriptive statistics for all five domains, including the \textit{Adjectives} to \textit{Non-adjectives} ratio and the median probability ($\theta_{domain}$) of the $t_{TC}(d)$ topic for each domain. As can be seen in this table, there is a significant number of \textit{Adjectives} in each domain, but the variance in their number is substantial. Also, \textit{Topics} are domain specific, with the most correlated topic $t_{TC}(d)$ for each domain being substantially more visible in its domain compared with the others. In Table \ref{tab:topic_words} we provide the top words for all \textit{Topics}, to show how they capture domain specific information.

\begin{table}[htbp]
    \centering
    \begin{tabular}{p{1.5cm}p{1.0cm}p{1.0cm}p{1.0cm}p{1.2cm}p{0.7cm}p{0.7cm}p{0.7cm}p{0.7cm}p{0.7cm}} \hline
    Domain & Min. $r(adj)$ & Med. \# $r(adj)$ & Max. \# $r(adj)$ & $\sigma$ of \# $r(adj)$ & $\theta_{b}$ & $\theta_{d}$ & $\theta_{e}$ & $\theta_{k}$ & $\theta_{m}$ \\ \hline
    Books & $0.0$ & $0.135$ & $0.444$ & $0.042$ & $0.311$ & $0.011$ & $0.052$ & $0.026$ & $0.014$ \\
    DVD & $0.0$ & $0.138$ & $0.425$ & $0.042$ & $0.014$ & $0.045$ & $0.045$ & $0.016$ & $0.225$ \\
    Electronics & $0.0$ & $0.136$ & $0.461$ & $0.049$ & $0.010$ & $0.065$ & $0.080$ & $0.039$ & $0.003$ \\
    Kitchen & $0.0$ & $0.142$ & $0.5$ & $0.052$ & $0.007$ & $0.039$ & $0.075$ & $0.066$ & $0.002$ \\
    Movies & $0.0$ & $0.138$ & $0.666$ & $0.0333$ & $0.010$ & $0.007$ & $0.045$ & $0.016$ & $0.281$ \\ \hline
    \end{tabular}
    \caption{Descriptive statistics for the Sentiment Classification datasets. $r(adj)$ denotes the ratio of \textit{Adjectives} to \textit{Non-adjectives} in an example. $\theta_{domain}$ is the mean probability of the topic that is most observed in that domain which will also serve as our \textit{treated topic}. $b,d,e,k,m$ are abbreviations for Books, DVD, Electronics, Kitchen and Movies.}
    \label{tab:sentiment-data}
\end{table}

\begin{table}[htbp]
    \centering
    \small
    \scalebox{0.7}{\begin{tabular}{|c|l|l|l|l|l|l|l|l|l|l|} \hline
\# &  \multicolumn{10}{c|}{Top $10$ Words} \\ \hline
1 & set & box & wait & 20 & making & flat & worth & longer & disappoint & spend \\
\textcolor{orange}{2} & \textcolor{orange}{pan} & \textcolor{orange}{phone} & \textcolor{orange}{computer} & \textcolor{orange}{work} & \textcolor{orange}{does} & \textcolor{orange}{use} & \textcolor{orange}{still} &\textcolor{orange}{non} & \textcolor{orange}{battery} & \textcolor{orange}{problem} \\
3 & great & use & just & months & problem & bought & time & good & years & ago \\
4 & classic & stories & know & great & really & book & definitely & reading & writing & long \\
5 & item & dull & returned & expect & given & fit & did & ridiculous & run & matter \\
6 & kids & crap & turned & fun & children & making & point & needs & understand & truly \\
\textcolor{green}{7} & \textcolor{green}{dvd} & \textcolor{green}{version} & \textcolor{green}{video} & \textcolor{green}{player} & \textcolor{green}{original} & \textcolor{green}{screen} & \textcolor{green}{release} & \textcolor{green}{quality} & \textcolor{green}{features} & \textcolor{green}{cover} \\
8 & book & real & second & school & author & going & page & shows & past & light \\
9 & machine & software & uses & issues & using & help & problems & makes & device & bought \\
10 & mind & fine & despite & pages & author & lost & books & book & read & especially \\
11 & book & reading & information & read & quot & books & better & author & know & does \\
12 & just & did & know & ll & does & ve & think & got & times & work \\
\textcolor{purple}{13} & \textcolor{purple}{product} & \textcolor{purple}{buy} & \textcolor{purple}{amazon} & \textcolor{purple}{bought} & \textcolor{purple}{plastic} & \textcolor{purple}{did} & \textcolor{purple}{reviews} & \textcolor{purple}{cheap} & \textcolor{purple}{work} & \textcolor{purple}{ve} \\
14 & does & man & just & woman & story & women & way & stop & time & like \\
15 & expected & star & series & rest & terrible & simply & pretty & watching & paid & wait \\
16 & away & water & stay & model & dog & good & difficult & like & right & just \\
17 & broke & replacement & warranty & month & send & weeks & days & called & week & product \\
18 & people & god & says & mr & life & like & world & person & american & way \\
19 & return & garbage & single & different & unless & given & oh & hot & plastic & thought \\
20 & play & does & power & light & white & little & used & make & drive & large \\
21 & bad & good & pretty & really & ve & just & worst & seen & 10 & best \\
\textcolor{red}{22} & \textcolor{red}{movie} & \textcolor{red}{film} & \textcolor{red}{like} & \textcolor{red}{movies} & \textcolor{red}{acting} & \textcolor{red}{bad} & \textcolor{red}{watch} & \textcolor{red}{just} & \textcolor{red}{plot} & \textcolor{red}{scenes} \\
23 & fan & wrote & fans & years & special & true & humor & day & disappoint & novel \\
24 & order & received & monster & performance & ordered & sent & said & better & later & returned \\
25 & book & long & ll & just & tell & totally & later & reader & given & great \\
26 & book & job & person & poor & read & kept & thought & trying & boring & good \\
27 & new & piece & tried & stopped & junk & worked & working & work & brand & maybe \\
28 & line & john & coming & certainly & early & true & films & enjoy & like & write \\
29 & book & read & books & author & pages & novel & writing & reader & history & interesting \\
30 & killer & card & camera & car & shows & stupid & series & tv & picture & better \\
31 & coffee & mouse & stand & products & use & like & make & decided & finally & tried \\
32 & john & writing & movie & book & waste & time & plot & make & did & line \\
33 & quot & written & self & does & things & view & needs & like & new & hope \\
34 & book & let & good & make & did & interesting & does & say & self & great \\
35 & unit & device & purchased & features & works & house & returned & running & warranty & hear \\
36 & does & hand & nice & need & small & clean & time & sex & look & things \\
37 & quality & poor & daughter & cable & low & design & control & sound & bad & good \\
\textcolor{blue}{38} & \textcolor{blue}{boring} & \textcolor{blue}{long} & \textcolor{blue}{time} & \textcolor{blue}{end} & \textcolor{blue}{story} & \textcolor{blue}{rest} & \textcolor{blue}{stop} & \textcolor{blue}{slow} & \textcolor{blue}{minutes} & \textcolor{blue}{good} \\
39 & old & year & horrible & great & got & food & beautiful & boy & said & instead \\
40 & hard & happy & sure & disappoint & writing & music & bad & reviews & days & uses \\
41 & known & christian & truth & like & feel & store & novel & remember & stay & able \\
42 & mouse & design & 15 & agree & purchased & given & job & happened & order & making \\
43 & world & war & words & self & old & word & attempt & needed & title & life \\
44 & lost & christian & guys & despite & turn & getting & mind & decent & war & fine \\
45 & music & ipod & weak & car & 30 & battery & playing & takes & able & major \\
46 & like & just & really & did & characters & story & character & love & little & make \\
47 & money & waste & time & save & thought & worth & spend & better & good & just \\
48 & disappointed & feel & fast & little & bit & good & job & parts & matter & complete \\
49 & day & black & sound & hours & like & just & minutes & bread & went & getting \\
50 & service & support & customer & told & product & check & company & called & terrible & hold \\  \hline
    \end{tabular}}
    \caption{Top 10 words in each of the 50 topics. A topic model was trained on all texts in all domains combined. Topic $\# 22$, our $\theta_{m}$, is highlighted in \textcolor{red}{red}, topic $\# 38$, $\theta_{b}$, is highlighted in \textcolor{blue}{blue}, topic $\# 8$, $\theta_{d}$, is highlighted in \textcolor{green}{green}, topic $\# 2$, $\theta_{e}$, is highlighted in \textcolor{orange}{orange} and topic $\# 13$, $\theta_{k}$, is highlighted in \textcolor{purple}{purple}. $b,d,e,k,m$ are abbreviations for the Books, DVD, Electronics, Kitchen and Movies domains.}
    \label{tab:topic_words}
\end{table}

Our sentiment classification data allows for a natural setting for testing our methods and hypotheses, but it has some limitations. Specifically, in the case of \textit{Topics}, we cannot generate realistic counterfactual examples and therefore compute $ATE_{gt}$, the ground-truth estimator of the causal effect. This is because creating counterfactual examples would require deleting the topic from the text without affecting the grammaticality of the text, something which cannot be done automatically. In the case of \textit{Adjectives}, we are hoping that removing \textit{Adjectives} will not affect the grammaticality of the original text, but are aware that this sometimes might not be the case. While this data provides a real-world example of natural language, it is hard to automatically generate counterfactuals for it. To allow for a more accurate estimation of the ground truth effect, we would need a dataset where we can control the data-generating process.

\subsection{The Enriched Equity Evaluation Corpus (EEEC)} 
\label{subsec:eeec}
Understanding and reducing gender and racial bias encapsulated in classifiers is a core task in the growing literature of interpretability and debiasing in NLP (see Section \ref{sec:prev}). There is an ongoing effort to both detect such bias and to mitigate its effect, which we see from a causal perspective as a call for action. By offering a way to estimate the causal effect of the \textit{Gender} and \textit{Race} concepts as they appear in the text, on classifiers, we enable researchers to avoid using biased classifiers. 

In order to evaluate the quality of our causal effect estimation method, we need a dataset where we can control test examples such that for each text we have a counterfactual text that differs only by the \textit{Gender} or \textit{Race} of the person it discusses. We also need to be able to control the data-generating process in the training set, so that we can create such a bias for the model to pick up. A dataset that offers such control exists, and is called the Equity Evaluation Corpus (EEC)~\cite{kiritchenko2018examining}.

It is a benchmark dataset, designed for examining inappropriate biases in system predictions, and it consists of $8,640$ English sentences chosen to tease out \textit{Racial} and \textit{Gender} related bias. Each sentence is labeled for the mood state it conveys, a task also known as \textit{Profile of Mood States} (POMS). Each of the sentences in the dataset is comprised using one of eleven templates, with placeholders for a person's name and the emotion it conveys. For example, one of the original templates is: \textit{"<Person> feels <emotional state word>."}. The name placeholder (\textit{<Person>}) is then filled using a pre-existing list of common names that are tagged as male or female, and as African-american or European.\footnote{In this paper we take a binary approach towards race and gender, as is done in the Equity Evaluation Corpus \cite{kiritchenko2018examining}, although this is obviously not the case in reality. This helps us keep the task and experiments clear, easy to follow and analyse.} The emotion placeholder (\textit{<emotional state word>}) is filled using lists of words, each list corresponding to one of four possible mood states: \textit{Anger}, \textit{Sadness}, \textit{Fear} and \textit{Joy}. The label is the title of the list from which the emotion is taken.

Designed as a bias detection benchmark, the sentences in EEC are very concise, which can make them not useful as training examples. If a classifier sees in training only a small number of examples, which differ only by the name of the person and the emotion word, it could easily memorize a mapping between emotion words and labels, and will not learn anything else. To solve this and create a more representative and natural dataset for training, we expand the EEC dataset,  creating an enriched dataset which we denote as \textit{Enriched Equity Evaluation Corpus}, or EEEC. In this dataset, we use the $11$ templates of EEC and randomly add a prefix or suffix phrase, which can describe a related place, family member, time and day, including also the corresponding pronouns to the \textit{Gender} of the person being discussed. We also create $13$ non-informative sentences, and concatenate them before or after the template such that there is a correlation between each label and $3$ of those sentences.\footnote{Each of those three sentences are five times more likely to appear than the other ten for that label.} This is performed so that we have other information that could be valuable for the classifier other than the person's name and the emotion word. Also, to further prevent memorization, we include emotion words that are ambiguous and can describe multiple mood states.

Our enriched dataset consists of $33,738$ sentences generated by $42$ templates that are longer and much more diverse than the templates used in the original EEC. While still synthetic and somewhat unrealistic, our datatest has much longer sentences, has more features that are predictive of the label and is harder for the classifier to memorize. In Appendix \ref{data-tables} we provide additional details about the EEEC dataset, through two tables: One that presents the templates used to generate the data, and one that compares the original EEC to our EEEC, illustrating the key modifications we have made.


For each example in EEEC we generate two counterfactual examples: One for \textit{Gender} and  one for \textit{Race}. That is, we create two instances which are identical except for that specific concept. For the \textit{Gender} case, we change the name and the \textit{Gender} pronouns in the example and switch them, such that for the original example: "\textit{Sara feels excited as she walks to the gym}" we will have the counterfactual example: "\textit{Dan feels excited as he walks to the gym}". For the \textit{Race} concept, we create counterfactuals such that for the same original example, the counterfactual example is: "\textit{Nia feels excited as she walks to the gym}". For each counterfactual example, the person's name is taken at random from the pre-existing list corresponding to its type.

%% file: chapters/tasks.tex
\section{Tasks and Experiments}
\label{sec:tasks}

Equipped with datasets for both Sentiment Classification and Profile of Mood States (POMS), and annotated for concepts (\textit{Adjectives}, \textit{Topics}, \textit{Gender} and \textit{Race}), we now define tasks and experiments for which we train classification models and test our proposed method for causal effect estimation of chosen concepts. In three of those cases (\textit{Adjectives}, \textit{Gender} and \textit{Race}) we have some control over the data-generating process, and therefore can compare the estimated causal effect to the ground truth effect. We start with experiments designed to estimate the effect of two concepts, \textit{Adjectives} and \textit{Topics}, on sentiment classification. We choose these concepts as representatives of local (\textit{Adjectives}, expressed as individual words or short phrases) and global (\textit{Topics}, expressed as distribution over the vocabulary) concepts that are intuitively related to sentiment analysis. Then, we explore the potential role of gender and racial bias in mood state classification. For each concept, we experiment with three versions of the data: \textit{Balanced}, \textit{Gentle} and \textit{Aggressive}, which differ by the correlation between the \textit{treated concept} and the label. In Table \ref{tab:tasks}, we summarize the four \textit{treated concepts} we experiment with. Table \ref{tab:experiments} presents the differences between the experiments we conduct for each \textit{treated concept} in terms of the concept-label correlation. 

\begin{table}[!htbp]
    \centering
    \small
    \begin{tabularx}{\textwidth}{|X|X|X|X|X|} \hline
        Concept & Task & Adversarial Task & Optional Control Tasks & Dataset  \\ \hline
        Adjectives & Sentiment & Masked Adjectives & PoS Tagging & Movie Reviews \\ \hline
        Topics & Sentiment  & Above Average Topic Prob. & Topic Class. & All Reviews \\ \hline
        Gender & POMS & Gender Class. & Race Class. & Enriched EEC \\ \hline
        Race & POMS & Race Class. & Gender Class. & Enriched EEC \\ \hline
    \end{tabularx}
    \caption{Summary of the tasks we experiment with. PoS stands for Part of Speech, POMS for Profile of Mood States and EEC for the Equity Evaluation Corpus. For each of the four tasks, we describe the task designed in order to forget the concept, alongside tasks designed to control against forgetting potential confounders.}
    \label{tab:tasks}
\end{table}

\begin{table}[!htbp]
    \centering
    \begin{tabular}{|l|l|l|l|l|} \hline
        \multirow{2}{*}{Treated Concept} & \multirow{2}{*}{Label} & \multicolumn{3}{|c|}{Concept-Label Correlation} \\ 
        & & Balanced & Gentle & Aggressive \\ \hline
        Adjectives & Sentiment & 0.056 & 0.4 & 0.76 \\
        Topic & Sentiment & 0.002 & 0.046 & 0.127 \\
        Gender & POMS & 0.001 & 0.074 & 0.245 \\
        Race & POMS & 0.005 & 0.069 & 0.242 \\ \hline
    \end{tabular}
    \caption{The correlation between the \textit{treated concept} and the label for each experiment we run (Balanced, Gentle and Aggressive). For each experiment we present the correlation on the full dataset (train, dev and test combined).}
    \label{tab:experiments}
\end{table}

With this experimental setup we seek to answer four research questions:
\begin{enumerate}
    \item Can we accurately approximate \textit{$ATE_{gt}$}, the ground-truth estimator of the causal effect, using our proposed \textit{TReATE} estimator ?
    \item Does \textit{BERT-CF}, our counterfactual representation model, forget the \textit{treated concept}?
    \item Does \textit{BERT-CF} remember the \textit{control concept}?
    \item Can \textit{BERT-CF} help mitigate potential bias in the downstream classifier ?
\end{enumerate}

In answering these questions, we hope to show that our method can provide accurate causal explanations that can be used in a variety of settings. Question $\#1$ is our core causal estimation question, where we wish to test whether the ground truth \textit{ATE} can be approximated with \textit{TReATE}. Questions $\#2$ and $\#3$ are important because we would like to know that the estimated effect we see in question $\#1$ is a result of our Stage 2 intervention that created \textit{BERT-CF} (see Figure \ref{fig:BERT-CF}), and not due to other reasons. Unlike question $\#1$, questions $\#2$ and $\#3$ do not require access to counterfactual examples, and can be used to validate our method in real-world settings. Finally, a byproduct of our method is \textit{BERT-CF}, a counterfactual representation model that is unaffected by the \textit{treated concept}. In question $\#4$ we ask if such a representation model can be useful in mitigating the perhaps unwanted effect of the \textit{treated concept} on the task classifier.

To tackle these questions, we start by describing how to estimate the causal effect for each of the \textit{treated concepts} while considering the potentially confounding \textit{control concepts} (question $\#1$). For each \textit{treated concept}, we explain how we control the concept-label correlation to create the \textit{Balanced}, \textit{Gentle} and \textit{Aggressive} versions. We then discuss how to answer questions $\#2$ and $\#3$ for a given \textit{TC} and \textit{CC}, and briefly explain how we answer question $\#4$ in the \textit{Aggressive} version. We detail our experimental pipeline and hyper-parameters in Appendix \ref{app:params}.

\subsection{The Causal Effect of Adjectives on Sentiment Classification}
\label{sec:adjectives}

Following the example we discuss in Section \ref{sec:intro}, we choose to measure the effect of \textit{Adjectives} on sentiment classification. In using \textit{Adjectives} as our \textit{treated concept}, we follow the discussion in the sentiment classification literature that marks them as linguistic features with a prominent effect. Another key characteristic of \textit{Adjectives} is that they can usually be removed from a sentence without affecting the grammaticality of the sentence and its coherence. Finally, with the recent advancement of parts-of-speech (\textit{PoS}) taggers~\cite{akbik2018coling}, we can rely on automatic models to tag our dataset with high accuracy, thus avoiding the need for manual tagging.

\begin{figure}[!htbp]
 \centering
 \includegraphics[width=0.5\linewidth]{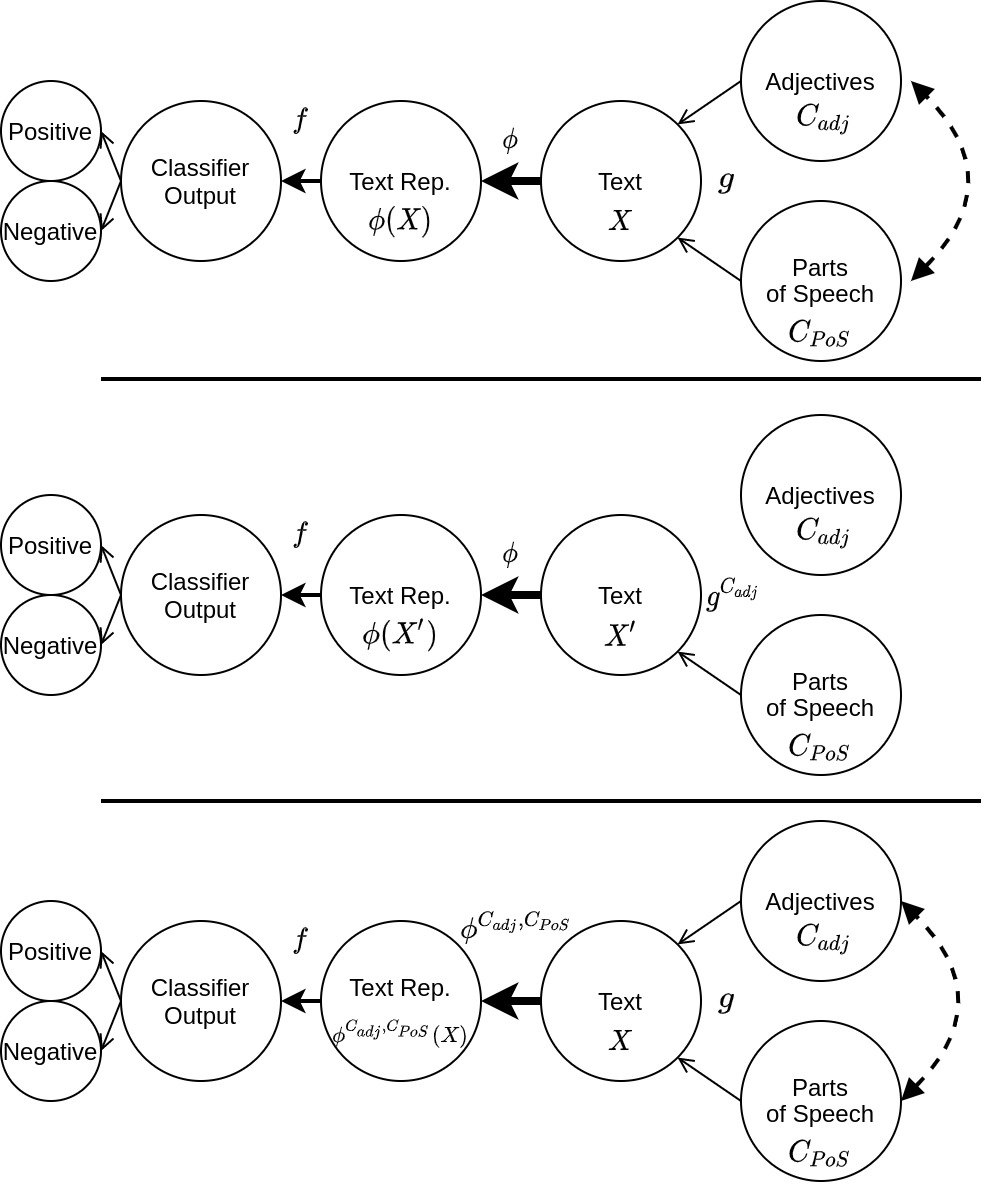}    
    \caption{A causal graph for \textit{Adjectives} and other \textit{Parts-of-Speech} generating a text with a positive or a negative sentiment. The top graph represents a data-generating process where all \textit{Parts-of-Speech} generate the texts, with a potential hidden confounder affecting both $C_{adj}$, the \textit{Treated Concept}, and $C_{PoS}$, the \textit{Control Concept}. The middle graph represents the scenario where we can control the generation process and create a text that is not influenced by the \textit{Treated Concept}. The bottom graph represents our approach, where we  manipulate the text representation.}
    \label{fig:cg-adj}    
\end{figure}

The causal graph we use to guide our choice of the \textit{treated} and \textit{control concepts} is similar to that of our motivating example, and is illustrated in Figure \ref{fig:cg-adj}. In the Sentiment reviews dataset (presented in Section \ref{subsec:sentiment_reviews}), since there are no concepts such as \textit{Political Figure} being discussed, we use other \textit{PoS} tags (i.e., everything but \textit{Adjectives}) as our \textit{control concepts}.

\paragraph{Controlling the Concept-Label Correlation}
Using the reviews dataset, we create multiple datasets, differing by the correlation between the ratio of \textit{Adjectives} and the label. We split the original dataset into training, development and test sets following a $64\%,16\%,20\%$ ($37120$, $9280$, $11600$ sentences) split, respectively. Then, we create three versions of the data: \textit{Balanced}, \textit{Gentle} and \textit{Aggressive}. In the \textit{Balanced} version we employ all reviews regardless of the ratio of \textit{Adjectives} they contain, preserving the data-driven correlation between the concept (\textit{Adjectives}) and the label (sentiment class). In the \textit{Gentle} version, we sort sentences from the \textit{Balanced} version by the ratio of \textit{Adjectives} they contain (in descending order) and delete the top half of the list for the sentences that appear within negative reviews, thus creating a negative correlation between the ratio of \textit{Adjectives} and the negative label in the train, development and test sets. For the \textit{Aggressive} version we do the same, and also delete the bottom half of the list for the sentences that appear within positive reviews, resulting in a higher correlation between the ratio of \textit{Adjectives} and the positive labels (see Table \ref{tab:experiments}).

\paragraph{Modelling the treated concept (TC) and the control concept (CC)} 
We follow the causal graph presented in Figure \ref{fig:cg-adj} and implement the adversarial \textit{Is Masked Adjective} ($IMA$) as our \textit{treated concept} (TC) objective shown in Equation \ref{eq:loss}. The $IMA$ objective is very similar to the $MLM$ objective, and it utilizes the same \textit{[MASK]} token used in  $MLM$, which masks each token to be predicted. However, instead of predicting the masked word we predict whether or not it is an adjective. To accommodate the $IMA$ prediction objective for any given input text, we masked all \textit{Adjectives} in addition to an equal number of non-adjective words, to ensure we result with a balanced binary classification token-level task. We follow the same probabilities suggested for the $MLM$ task in~\citet{devlin2018bert}.\footnote{The probabilities used in the original BERT paper are: $80\%$, $10\%$ and $10\%$ for masking the token, keeping the original token or changing it to a random token, respectively.}

For the \textit{control concept} (CC) task, we utilize all \textit{PoS} tags apart from \textit{Adjectives}, and train a sequence tagger to classify each \textit{Non-adjective} word according to its \textit{PoS}.\footnote{To prevent the model from learning to associate the null label with \textit{Adjectives}, we do not add it to the loss.} This serves the purpose of preserving syntactic concepts other than \textit{Adjectives}. In Section \ref{sec:results} we discuss the effect of the \textit{CC} objective on our estimates. Finally, as explained in Section \ref{sec:meth} (see Equation \ref{eq:loss}), to produce the \textit{BERT-CF} model for \textit{Adjectives}, we adversarially train the $IMA$ objective jointly with the other terms of the objective that are trained in a standard manner.

\subsection{The Causal Effect of Topics on Sentiment Classification}
\label{sec:topics}

Another interesting concept that we can explore using the reviews dataset is \textit{Topics}, as captured by the Latent Dirichlet Allocation (LDA) model~\cite{blei2003latent}. \textit{Topics} capture high-level semantics of documents, and are widely used in NLP for many language understanding purposes~\cite{boyd2017applications, oved2019predicting}. \textit{Topics} are qualitatively different from \textit{Adjectives}, as \textit{Adjectives} are concrete and local while \textit{Topics} are abstract and global. In the context of sentiment classification, it is reasonable to assume that the \textit{Topic} being discussed has an effect on the probability of the review being positive or negative. For example, some movie genres generally get more negative reviews than others, and some products are more generally liked than their alternatives. A key advantage of \textit{Topics} for our purposes is that they can be trained without supervision, allowing us to test this concept without manual tagging.

\begin{figure}[!htbp]
 \centering
 \includegraphics[width=0.5\linewidth]{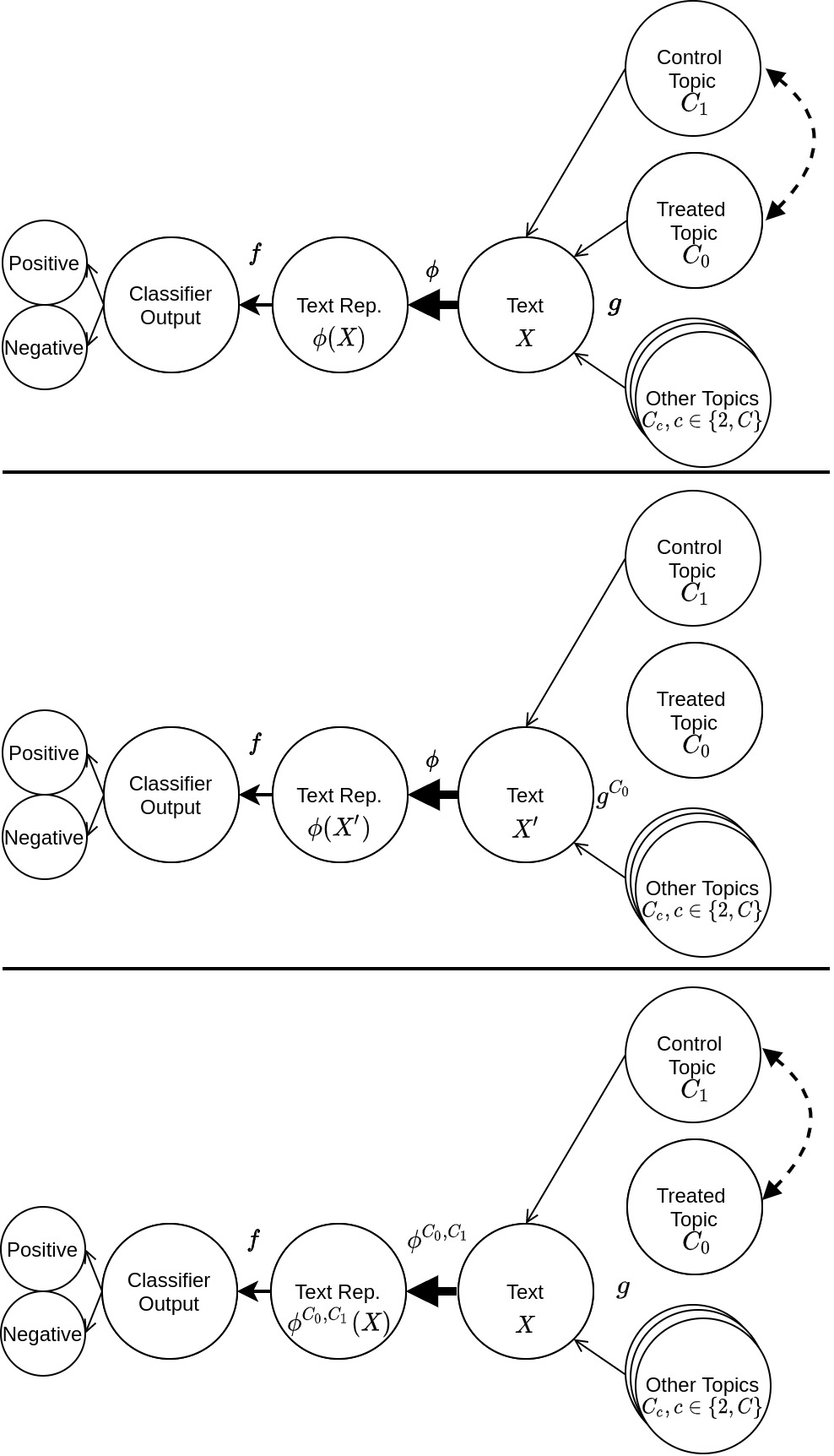}    
    \caption{A causal graph for \textit{Topics} generating a review with a positive or negative sentiment. The top graph represents a data-generating process where \textit{Topics} generate texts, with a potential hidden confounder affecting both $C_0$, the \textit{Treated Topic}, and $C_1$, the \textit{Control Topic}. The middle graph represents the scenario where we can control the generation process and create a text without the \textit{Treated Topic}. The bottom graph represents our approach, where we  manipulate the text representation.}
    \label{fig:cg-topics}    
\end{figure}

\textit{Topics} are global concepts that encode information across the different reviews in the corpus. Yet, by using topic modeling we can represent them as variables that come with a probability that reflects the extent to which they are represented in each document. This allows us to naturally integrate them into our \textit{TC} term presented in Figure \ref{fig:BERT-CF} (i.e the \textit{treated concept}), but also to the preserving \textit{CC} term (the \textit{control concept}). In Figure \ref{fig:cg-topics}, we illustrate the causal graph that we follow. For the \textit{treated} (\textit{TC}) and \textit{control} (\textit{CC}) \textit{Topics}, we follow Equation \ref{eq:topic} and use the \textit{Topics} $t_{TC}(domain)$ and $t_{CC}(domain)$, which we denote as $C_0$ and $C_1$, respectively.

Unlike the \textit{Adjectives} experiments, we can not directly manipulate the texts to create counterfactual examples for \textit{Topics}. For a given document, changing the topic being discussed cannot be done by simply deleting the words associated with it, and would require rewriting the text completely. As an alternative, we can use the domain variation in the reviews dataset and the correspondence of some \textit{Topics} to specific domains, to test the performance of our causal effect estimator, \textit{TReATE}. We see this as a unique contribution of this experiment as it allows us to test our causal effect estimator in a case where we do not have access to the ground-truth (estimation of the) causal effect.

Another issue with \textit{Topics} is that they are confounders for one another by design. LDA models texts as mixtures of \textit{Topics}, where each \textit{Topic} is a probability distribution over the vocabulary. As \textit{Topics} are on the simplex (they are a probability distribution), if the probability of one \textit{Topic} decreases, the probability of others must increase. For example, if the example presented in Section \ref{sec:intro} was less about politics, it would have to be more about a different \textit{Topic}. Below we show how we circumvent the effect of those potential confounders in our TC and CC objectives as shown in Equation \ref{eq:loss}.

\paragraph{Controlling the Concept-Label Correlation}
For the \textit{Topics} experiments, we also create three versions of the data, following a similar \textit{Balanced}, \textit{Gentle} and \textit{Aggressive} approaches and using the reviews data as above. For the \textit{Balanced} version, we use all of the data from \textit{Books}, \textit{DVD}, \textit{Electronics}, \textit{Kitchen Appliances} and \textit{Movies} domains. For the \textit{Gentle} version, we take the \textit{Balanced} dataset and delete half of the negative reviews where the  $t_{TC}(Movies)$ topic is less represented (with probability lower than the median probability), resulting in a positive correlation between the topic and the positive label. For the \textit{Aggressive} version we also delete half of the positive reviews where the  $t_{TC}(Movies)$ topic is more represented, thus further increasing the correlation between the topic and the labels. For all these experiments we follow the same $64\%,16\%,20\%$ split for the training, development and test sets, respectively, as for the \textit{Adjectives} experiments. As another set of experiments, we follow the same steps for the \textit{Gentle} and \textit{Aggressive} versions where the chosen topic is $t_{TC}(Books)$ instead of $t_{TC}(Movies)$.

As we do not have access to real counterfactual examples in this case, we can only compute \textit{TReATE} for a given test-set and qualitatively analyze the results. Particularly, the multi-domain nature of our dataset allows us to estimate \textit{TReATE} on each domain separately, and test whether the estimated effect varies between domains. Specifically, we can test whether for a given $t_{TC}(domain)$ the $TReATE_{t_{TC}(domain)}$ estimator is higher on domains where $t_{TC}(domain)$ is more present, compared with those domains where it is less present. To do that, we compute the estimated \textit{TReATE} for each $t_{TC}(domain)$ (\textit{Books} and \textit{Movies}) on each of the five domains separately, and discuss the results in Section \ref{sec:results}. We focus most of the discussion on these experiments in Section \ref{subsec:results-analysis}, where we test whether we can successfully mitigate the bias introduced in the \textit{Gentle} and \textit{Aggressive} version.

\paragraph{Modelling the treated concept (TC) and the control concept (CC)} 
Using the binary variables indicating if for a given topic the probability is above or below its median (see Section \ref{sec:data}), we introduce "\textit{Is Treated Topic}" ($ITT$), a binary adversarial fine-tuning task for our \textit{treated concept} (TC). As the \textit{TC}, we choose the $t_{TC}(domain)$ topic introduced in Section \ref{sec:data} in Equation \ref{eq:topic}. To control for the potential forgetting of related \textit{Topics}, we add alongside the adversarial task the prediction of the second most correlated topic, $t_{CC}(domain)$, as our \textit{control concept}, and add it as another fine-tuning task which we name "\textit{Is Control Topic}" ($ICT$). Finally, as explained in Section \ref{sec:meth} (see Equation \ref{eq:loss}), to produce the \textit{BERT-CF} model for \textit{Topics}, we adversarially train the $ITT$ objective jointly with the other objective terms that are trained in a standard manner.

\subsection{The Causal Effect of Gender and Racial Bias}
\label{sec:poms}

\begin{figure}[!htbp]
 \centering
 \includegraphics[width=0.5\linewidth]{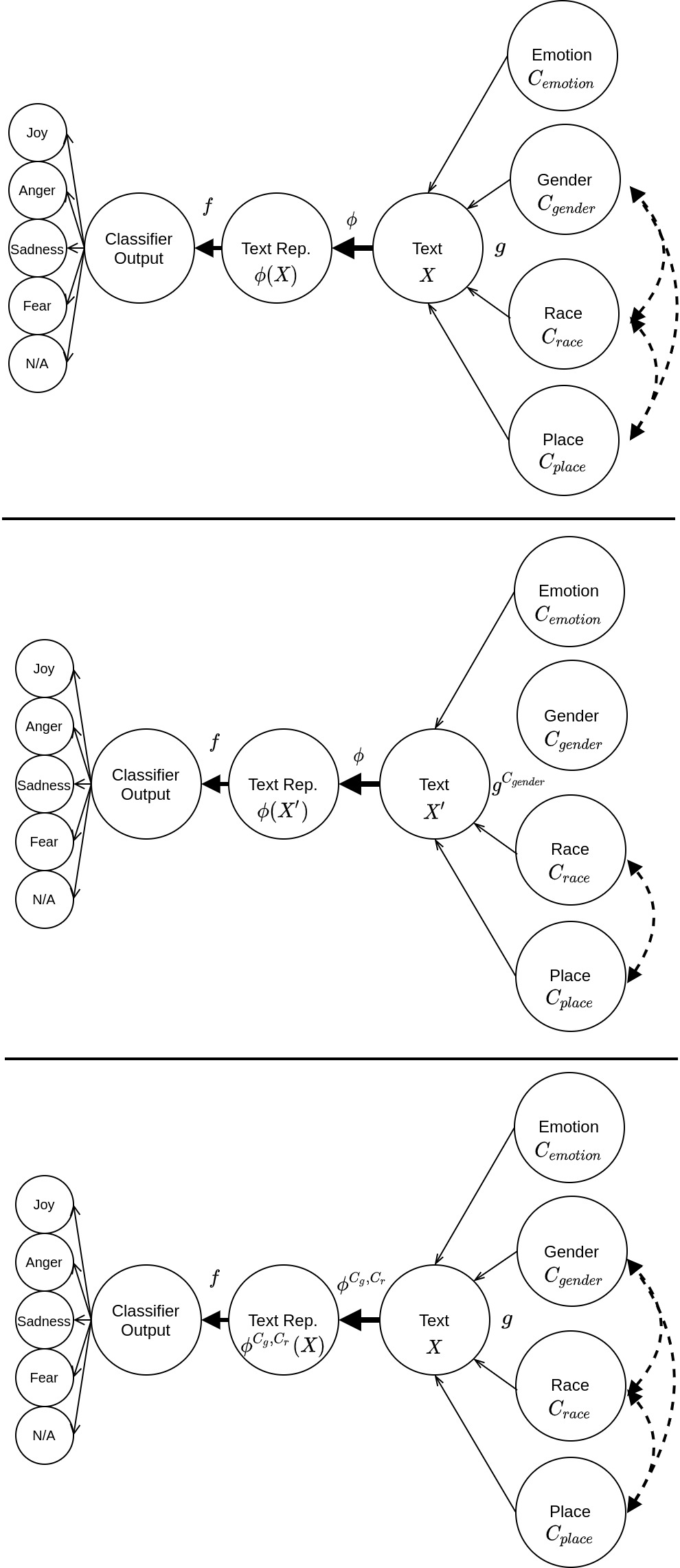}    
    \caption{A causal graph for \textit{Emotions}, \textit{Gender}, \textit{Race} and \textit{Place} generating a text with one of five mood states. The top graph represents a data-generating process where those concepts generate texts, with a potential hidden confounder affecting both $C_{gender}$, the \textit{treated concept}, and $C_{race}$, the \textit{control concept}. The middle graph represents the scenario where we can control the generative process and create a text without the \textit{treated concept}. The bottom graph represents our approach, where we manipulate the text representation.}
    \label{fig:cg-gender}    
\end{figure}

While \textit{Adjectives} and \textit{Topics} capture both local and global linguistic concepts, our ability to generate counterfactual examples for them is limited. Particularly, for \textit{Topics} we cannot generate counterfactual examples, while for \textit{Adjectives} we use real-world data and hence our control on the data generating process is limited. 
To allow for a more accurate comparison to the true causal effect, we consider two tasks, \textit{Gender} and \textit{Race}, where such a comparison can be made using the EEEC dataset presented in Section \ref{sec:data}. In Figure \ref{fig:cg-gender}, we illustrate the causal graph for the case where \textit{Gender} is the \textit{treated concept}. We denote \textit{Gender} as $C_{gender}$, our \textit{treated concept} (\textit{TC}), and the potentially confounding concept is $C_{race}$, our \textit{control concept} (\textit{CC}).
The \textit{Race} task is generated similarly, by simply replacing \textit{Gender} and \textit{Race} in the causal graph. 

As this dataset is constructed using the templates described in Table \ref{tab:eeec-templates}, we can directly control each concept and create a true counterfactual example for each sentence. For instance, we can take a sentence that describes a European male being angry, and replace his name (and the relevant pronouns) to a European female. Holding the \textit{Race} and the rest of the sentence fixed, we can measure the true causal effect as the difference in a model's class distribution on the original European male example compared to that of the counterfactual, European female example.

Another advantage of experimenting with \textit{Gender} and \textit{Race} is that their effect, if exists, is often undesirable. If we can use our method to create an unbiased textual representation with respect to the \textit{treated concept}, then we can create better, more robust models using this representation. In Section \ref{subsec:results-analysis} we discuss how to use our \textit{BERT-CF} representation to mitigate such bias and create better performing models.

\paragraph{Controlling the Concept-Label Correlation}
Using the EEEC data presented in Section \ref{sec:data}, we create multiple versions of the dataset, differing by the correlation between \textit{Gender}/\textit{Race} and the labels. For both \textit{Gender} and \textit{Race}, we create three versions of the data: \textit{Balanced}, \textit{Gentle} and \textit{Aggressive}. In the \textit{Balanced} version, we randomly choose the person's name, resulting in almost no correlation between each label and the concept. In the \textit{Gentle} version, we choose names such that $90\%$ of examples from the \textit{Joy} label are of female names, and $50\%$ of the \textit{Anger}, \textit{Sadness} and \textit{Fear} examples are of male names. The \textit{Aggressive} version is created similarly, but with $90\%$ for \textit{Joy} and $10\%$ for the rest. For all these experiments we follow the same $64\%,16\%,20\%$ split for the training, development and test sets, respectively.

\paragraph{Modelling the treated concept (TC) and the control concept (CC)} 
In the case of \textit{Gender} and \textit{Race}, in order to produce the \textit{BERT-CF} model, the TC and CC are rather straightforward. For a given \textit{TC}, for example \textit{Gender}, we define a binary classification task, where for each example the classifier predicts the gender described in the example. Equivalently, the \textit{CC} task is also a binary classification task where, given that \textit{Gender} is the \textit{TC}, the classifier for \textit{CC} predicts the \textit{Race} described in the example.

\subsection{Comparing Causal Estimates to the Ground Truth}
\label{sec:ground-truth-comparison}
While we do not usually have access to ground truth data (i.e., counterfactual examples), we can artificially generate such examples in some cases. For instance, in the \textit{Gender} and \textit{Race} cases we have created a dataset where for each example we manually created an instance which is identical except that the concept is switched in the text. Specifically, we can switch the gender of the person being mentioned, holding everything else equal. For \textit{Adjectives}, we followed a similar process of producing counterfactual examples, where \textit{Adjectives} were removed from the original example's text. With these datasets we can then estimate the causal concept effect using our method, and compare this estimation to the ground truth effect, i.e., the difference in output class probabilities between actual test set examples and their manually created counterfactuals. Our ground-truth estimator of the causal effect is then an estimator of the \textit{Averaged Treatment Effect} (ATE, Equation \ref{eq:ATE}): 
\begin{equation}\label{eq:ATE_hat}
    ATE_{gt}(O) = \frac{1}{|I|} \left[ \sum_{i \in I}\langle \vec{z}(f(\phi^{O}(x_{i,C_0=1}))) - \vec{z}(f(\phi^{O}(x_{i,C_0=0}))) \rangle \right]
\end{equation}
Where $x_{i,C_0=1}$ is an example where the concept $C_0$ takes the value of 1, and $x_{i,C_0=0}$ is the same example, except that $C_0$ takes the value of 0. For instance, if $x_{i,C_0=1}$ is: \textit{"A woman is walking towards her son"}, $x_{i,C_0=0}$ will be: \textit{"A man is walking towards his son"}. Finally, 
$\vec{z}(\cdot)$ is the vector of output class probabilities assigned by the classifier when trained with $\phi^O$, the representation of a vanilla, unmanipulated pre-trained BERT model (denoted with \textit{BERT-O}, see below; to simplify our notation, we refer to this model simply as $O$).

\paragraph{Correlation-based Baselines}
We compare our methods to two correlation-based baselines, which do not take into account counterfactual representations and simply compute differences in predictions between test examples that contain the concept (i.e., $C_{TC}=1$) and those that do not ($C_{TC}=0$). The first baseline we consider is called \textit{CONEXP}, and it was proposed by \citet{goyal2019explaining} as an alternative for measuring the effect of a concept on models' predictions.  \textit{CONEXP} computes the conditional expectation of the prediction scores conditioned on whether or not the concept appears in the text. Importantly, this baseline is based on passive observations and is not based on \emph{do}-operator style interventions. The corpus-based estimator of \textit{CONEXP} is defined as follows:

\begin{equation}\label{eq:conexp}
\text{CONEXP}_{C_0}(O) = \langle \frac{1}{|I_{C_j=1}|} \sum_{i \in I_{C_j=1}} \vec{z}(f(\phi^O(x_{i})) - \frac{1}{|I_{C_j=0}|} \sum_{i \in I_{C_j=0}} \vec{z}(f(\phi^O(x_{i})) \rangle 
\end{equation}

The second baseline we consider is  \textit{TPR-GAP}, introduced in~\citet{de2019bias} and used by~\citet{ravfogel2020null}. \textit{TPR-GAP} computes the difference between the fraction of correct predictions when the concept exists in the text, and fraction of correct predictions when the concept does not exist in the text. It is computed using the following equation:
\begin{equation}\label{eq:tpr}
    \text{TPR-GAP}_{C_0}(O) = \sum_{l \in L} | P\big(f(\phi^O(X))=l|C_0=1,Y=l) - P\big(f(\phi^O(X)) = l|C_0=0,Y=l \big) |
\end{equation}
Where $P$ is the share of accurate model predictions, and $f(\phi^O(X))$ and $l \in L$ denote the predicted class and the correct class, respectively. 

Unlike \textit{CONEXP}, \textit{TPR-GAP} compares the accuracy of the model in two conditions, and not its class probability distribution, which prevents us from directly comparing it to the ground-truth  \textit{$ATE_{gt}(O)$} or to our \textit{TReATE}. As direct comparisons are not feasible, we discuss in Section \ref{sec:results} how the \textit{TPR-GAP} captures the concept effect compared with our \textit{TReATE}.

\paragraph{Language Representations}

In our experiments, we consider three different language representations, that are then used in the computations of our \textit{TReATE} causal effect estimator (Equation \ref{eq:TReATE_CC}), and the ground truth \textit{ATE} (Equation \ref{eq:ATE}):
\begin{itemize}
    \item \textit{BERT-O} - The representation taken from a pre-trained BERT, without any manipulations.
    \item \textit{BERT-MLM} - The representation from a BERT that was further fine-tuned on our dataset.
    \item \textit{BERT-CF} - The representation from BERT following our Stage 2 intervention (See Equation \ref{eq:loss} and Figure \ref{fig:BERT-CF}).
\end{itemize}

Recall that our experiments are designed to compare the predictions of BERT-based classifiers. For each experiment on each task, we compare for each test-set example the predictions of three trained classifiers, differing by the representations they use as input. To compute the estimator of the ground-truth causal effect, $ATE_{gt}(O)$, we compare the prediction of the \textit{BERT-O} based model on the original example to its prediction on the counterfactual and average on the entire test-set. Put it formally, we compute $ATE_{gt}(O)$ with Equation \ref{eq:ATE_hat} where $f$ is the \textit{BERT-O} based classification model. For our estimation of \textit{TReATE}, we compare for each example the prediction of the \textit{BERT-O} based model on the original example to the prediction of the \textit{BERT-CF} based model on the same example. 

As we want to directly evaluate the effect of our counterfactual training method, we also compute $\text{TReATE}(O,MLM)$. This estimator is equivalent to Equation \ref{eq:TReATE_hat} except that the \textit{BERT-CF} based classifier is replaced with a classifier that is based on \textit{BERT-MLM}: A representation model that is fine-tuned on the same data as \textit{BERT-CF}, but using the standard $MLM$ task instead of counterfactual training. Explicitly, we compute \textit{TReATE} using the following equation:

\begin{equation}\label{eq:TReATE_hat_O_CF}
    \text{TReATE}(O,CF) = \frac{1}{|I|} \left[ \sum_{i \in I} \langle \vec{z}(f(\phi^{O}(x_{i}))) - \vec{z}(f(\phi^{CF}(x_{i}))) \rangle \right]
\end{equation}
$\text{TReATE}(O,MLM)$ is computed using the same equation where $\phi^{CF}$ is replaced with $\phi^{MLM}$.

%% file: chapters/results.tex
\section{Results}
\label{sec:results}

Examining and analyzing our results, we wish to address the four research questions posed in Section~\ref{sec:tasks}. That is, we assess whether our method can accurately estimate the \textit{ATE} when such ground truth exists (question $\#1$), whether our \textit{BERT-CF} forgets the \textit{treated concept} and remembers the \textit{control concept} (questions $\#2$ and $\#3$, respectively) and whether we can mitigate bias using the \textit{BERT-CF} (question $\#4$). Finally, we dive into to the training process, and discuss the effect of our Stage 2 intervention on BERT's loss function.

\subsection{Estimating TReATE (The Causal Effect)}

\paragraph{Comparing TReATE and the Ground Truth ATE}

Our estimated $\text{TReATE}(O,CF)$ for each of the three concepts we have ground truth data for (\textit{Adjectives}, \textit{Gender} and \textit{Race})), compared to the ground truth ($\text{ATE}_{gt}(O)$) and the $\text{CONEXP}(O)$ baseline, are described in Tables \ref{tab:results-sentiment} and \ref{tab:results-POMS}.\footnote{We have also computed results for $\text{CONEXP}(MLM)$, $\text{TReATE}(MLM,CF)$ and $\text{ATE}_{gt}(MLM)$, but do not discuss them here as they are very similar and therefore do not add insight to this discussion.} In the \textit{Gender} and \textit{Race} experiments, we also compare our results to those obtained by the Iterative Nullspace Projection ($\text{INLP}$) method \cite{ravfogel2020null}. This method removes information from neural representations, but it does not preserve the information in control concepts. We let INLP remove information about the treated concept from the \textit{BERT-O}'s representation, and compute the $\text{TReATE}(O,INLP)$ using their default classification algorithm.\footnote{We utilize the original code from the author's GitHub repository, with its default hyperparameters: \url{https://github.com/shauli-ravfogel/nullspace_projection}.}

As demonstrated in the tables, we can successfully estimate the $\text{ATE}_{gt}(O)$ using our proposed $\text{TReATE}(O,CF)$: The values of $\text{TReATE}(O,CF)$ and $\text{ATE}_{gt}(O)$ are very similar across all experiments. Regardless of the amount of bias introduced in the experiments (\textit{Balanced}, \textit{Gentle} and \textit{Aggressive}), our method can estimate the causal effect successfully. Comparatively, the non-causal baseline $\text{CONEXP}(O)$ substantially underestimates the concepts' effect in $7$ out of $9$ experiments. In the other two experiments, the \textit{Balanced} and \textit{Gentle} \textit{Race} experiments, it overestimates the effect. Estimating $\text{TReATE}(O,CF)$ with INLP (referred to as $\text{TReATE}(O,INLP)$  above and as INLP in the table) substantially overestimates the $\text{ATE}_{gt}(O)$, possibly because INLP does not preserve the information encoded about control concepts.

In the \textit{Adjectives} experiments (Table \ref{tab:results-sentiment}) we see that the effect of \textit{Adjectives} on sentiment classification is prominent even in the \textit{Balanced} setting, suggesting that \textit{Adjectives} change the classifier's output class probability distribution by $0.397$ on average. While the bias introduced in the \textit{Gentle} setting did not affect the degree to which the classifier relies on \textit{Adjectives} in its predictions ($\text{ATE}_{gt}(O) = 0.397$ in the \textit{Balanced} case and $\text{ATE}_{gt}(O) = 0.376$ in the \textit{Gentle} case), it certainly did in the \textit{Aggressive} setting ($\text{ATE}_{gt}(O) = 0.634$ in the \textit{Aggressive} case). Interestingly, the effect of \textit{Adjectives} on the classifier is similar in the \textit{Balanced} and \textit{Gentle} settings, suggesting that the model was not fooled by the weak correlation between the number of \textit{Adjectives} and the positive label. When this correlation is increased, as was done in the \textit{Aggressive} setting, the effect increases by $60\%$ (from $\text{ATE}_{gt}(O) = 0.397$ in the \textit{Balanced} case to $\text{ATE}_{gt}(O) = 0.634$ in the \textit{Aggressive} case). 

\begin{table}[!htbp]
    \centering
    \begin{tabular}{l|c|c|c} 
    \toprule
    Experiment & $\text{ATE}_{gt}(O)$ & $\text{TReATE}(O,CF)$ & $\text{CONEXP}(O)$ \\ 
    \midrule
    Balanced & $0.397$ & $0.385$ & $0.01$ \\ 
    $[CI]$ & $[0.377, 0.417]$ & $[0.381, 0.389]$ & $[0,0.044]$ \\
    Gentle & $0.376$ & $0.351$ & $0.094$ \\
    $[CI]$ & $[0.361, 0.392]$ & $[0.347, 0.355]$ & $[0.061,0.127]$ \\
    Aggressive & $0.634$ & $0.603$ & $0.126$ \\ 
    $[CI]$ & $[0.613, 0.655]$ & $[0.588, 0.618]$ & $[0.095,0.158]$ \\
    \bottomrule
    \end{tabular}
    \caption{Results for the causal effect of \textit{Adjectives} on sentiment classification on Reviews. We compare $\text{TReATE}(O,CF)$ to the ground truth $\text{ATE}_{gt}(O)$ and the baseline $\text{CONEXP}(O)$. Confidence intervals ($[CI]$), computed using the standard deviations of $\text{ITE}_{gt}(O)$, $\text{TReITE}(O,CF)$ and $\text{CONEXP}$, are provided in square brackets.}
    \label{tab:results-sentiment}
\end{table}

\begin{table}[!htbp]
    \centering
    \begin{tabular}{l|l|c|c|c|c}
    \toprule
    \textit{TC} & Experiment & $ATE_{gt}$ & $\text{TReATE}$ & $\text{CONEXP}$ & $\text{INLP}$ \\ 
    \midrule
    \multirow {6}{*}{\textit{Gender}} & Balanced & $0.086$ & $0.125$ & $0.02$  & $0.313$ \\ 
     & $[CI]$ & $[0.082, 0.09]$ & $[0.110, 0.14]$ & $[0.0,0.05]$  & $[0.304, 0.321]$ \\ 
     & Gentle & $0.113$ & $0.135$ & $0.076$  &  $0.331$ \\ 
     & $[CI]$ & $[0.108, 0.118]$ & $[0.12, 0.15]$ & $[0.072,0.08]$  & $[0.322, 0.340]$ \\ 
     & Aggressive & $0.210$ & $0.241$ & $0.057$  & $0.437$ \\ 
     & $[CI]$ & $[0.203, 0.217]$ & $[0.229, 0.253]$ & $[0.051,0.063]$  & $[0.427, 0.448]$ \\ \midrule \midrule
     \multirow {6}{*}{\textit{Race}} & Balanced &  $0.014$ & $0.046$ & $0.08$  & $0.591$ \\ 
     & $[CI]$ &  $[0.012, 0.016]$ & $[0.038, 0.054]$ &  $[0.02, 0.014]$  & $[0.578, 0.605]$ \\ 
     & Gentle &  $0.027$ & $0.04$ & $0.044$  & $0.621$ \\ 
     & $[CI]$ &  $[0.024, 0.03]$ & $[0.048, 0.032]$ &  $[0.028, 0.07]$  & $[0.607, 0.635]$ \\ 
     & Aggressive &  $0.345$ & $0.332$ & $0.19$  & $0.650$ \\ 
     & $[CI]$ & $[0.333, 0.357]$ & $[0.324, 0.34]$ &  $[0.08, 0.3]$  &  $[0.636, 0.664]$ \\ 
    \bottomrule
    \end{tabular}
    \caption{Results for the effect of \textit{Gender} (top) and \textit{Race} (bottom) on POMS classification with the EEEC dataset. We compare $\text{TReATE}(O,CF)$ (the \text{TReATE} column) to the ground truth $\text{ATE}_{gt}(O)$ and the baseline $\text{CONEXP}(O)$. We also compare to INLP: A $\text{TReATE}(O,INLP)$ estimator where $\text{TReATE}(O,CF)$ is estimated with INLP instead of our BERT-CF.  Confidence intervals ($[CI]$), computed using the standard deviations of $\text{ITE}_{gt}(O)$, $\text{TReITE}(O,CF)$ $\text{CONEXP}$ and $\text{TReATE}(O,INLP)$  are provided in square brackets.}
    \label{tab:results-POMS}
\end{table}

In all three settings of the \textit{Adjectives} experiments, our $\text{TReATE}(O,CF)$ estimator is very similar to the $\text{ATE}_{gt}(O)$, and the gap between the two remains at $3\%$ (absolute). Similar patterns can be observed in the \textit{Gender} and \textit{Race} experiments(Table \ref{tab:results-POMS}). For both the \textit{Gender} and \textit{Race} concepts, we successfully approximate the $\text{ATE}_{gt}(O)$ with our $\text{TReATE}(O,CF)$ with a maximal error of $3.9\%$ (absolute) and an average error of $2.6\%$ (absolute). Similar to our observation in the \textit{Adjectives} case, in the \textit{Gender} and \textit{Race} cases the effect of the \textit{Gentle} bias on the extent to which the classifier relies on the \textit{treated concept} is very small. For both \textit{Gender} and \textit{Race}, the effect in the \textit{Gentle} setting is only slightly higher than that observed in the \textit{Balanced} setting ($1\%$ and $1.3\%$ absolute increase in $\text{ATE}_{gt}(O)$).

Another interesting pattern that emerges, is that the effect of \textit{Gender} on the POMS classifier in the \textit{Balanced} setting is $0.086$, more than six times higher than the $0.014$ observed in the equivalent \textit{Race} experiment. In our EEEC dataset, the \textit{Balanced} setting is designed such that there is no correlation between the \textit{Gender} or the \textit{Race} of the person and the label. The fact that such causal effect is observed suggests that \textit{BERT-O} contains \textit{Gender}-related information that affects classification decisions on downstream tasks.

To conclude, comparing $\text{TReATE}(O,CF)$ and $\text{ATE}_{gt}(O)$ on all experiments where we have counterfactual examples, we conclude that we can successfully estimate the causal effect, answering question $\#1$ presented in Section \ref{sec:tasks}. Regardless of the bias introduced and the extent that it affects the classifier, our $\text{TReATE}(O,CF)$ estimator remains close to the $\text{ATE}_{gt}(O)$. It can successfully detect bias when it exists and performs well even when the true effect is close to $0$ such as in the \textit{Balanced} \textit{Race} experiment. Comparatively, the $\text{CONEXP}(O)$ baseline is not able to approximate the true causal effect in any of the experiments we conduct here.

While we cannot directly compare the $\text{TPR-GAP}(O)$ baseline to the $\text{TReATE}(O,CF)$, $\text{ATE}_{gt}(O)$ and $\text{CONEXP}(O)$ estimators (Section \ref{sec:ground-truth-comparison}), we can still analyse the values of this non-causal estimator on each of the three versions for each concept. As seen in Table \ref{tab:results-tpr}, the $\text{TPR-GAP}(O)$ values are very small, with an average of $0.025$. While the results shown in Tables \ref{tab:results-sentiment} and \ref{tab:results-POMS} suggest that the true effect ($\text{ATE}_{gt}(O)$) increases along with the bias that was introduced to the data (with the exception of the \textit{Balanced} and \textit{Gentle} \textit{Adjectives} experiments), it is not the case with $\text{TPR-GAP}(O)$. The estimated effect in the \textit{Gentle} experiments is the highest for both \textit{Adjectives} and \textit{Gender} ($0.074$ and $0.014$, respectively), and is lowest for the \textit{Aggressive} experiments ($0.012$ and $0.003$, respectively). Only in the \textit{Race} experiments we see a pattern that is similar to that observed in the $\text{TReATE}(O,CF)$ and the $\text{ATE}_{gt}(O)$, but the scale is different, with a $2$ and $12$ fold increases in the $\text{ATE}_{gt}(O)$ ($0.014$ to $0.027$ and then to $0.345$), compared with a $6$ and $4$ fold increases in the $\text{TPR-GAP}(O)$ ($0.002$ to $0.012$ and then to $0.049$).

\begin{table}[!htbp]
    \centering
    \begin{tabular}{l|c|c|c} 
    \toprule
    Experiment & Adjectives & Gender & Race \\ 
    \midrule
    Balanced & $0.057$ & $0.003$ & $0.002$ \\ 
    Gentle & $0.074$ & $0.014$ & $0.012$ \\
    Aggressive & $0.012$ & $0.003$ & $0.049$ \\ 
    \bottomrule
    \end{tabular}
    \caption{The \textit{TPR-GAP} results for all versions (\textit{Balanced}, \textit{Gentle} and \textit{Aggressive}) for the three concepts where we have ground truth (\textit{Adjectives}, \textit{Gender} and \textit{Race}).}
    \label{tab:results-tpr}
\end{table}

\paragraph{Understanding TReATE Without Ground Truth}
Unlike the \textit{Adjdctives}, \textit{Gender} and \textit{Race} experiments, we do not have counterfactual examples for \textit{Topics} and therefore cannot compare our estimates to the ground truth. Alternatively, we provide here several sanity checks that suggest that the causal effect can be estimated for \textit{Topics} as well. With \textit{Topics} as our concepts we conduct two rounds of experiments, presented in Table \ref{tab:results-topics}, where in the first we choose $t_{TC}(books)$ and in the second we choose $t_{TC}(movies)$ as our \textit{treated concepts} (the \textit{control concepts} are chosen for each of \textit{TC} as described in Section \ref{sec:tasks}). For each, we train the models on the combined dataset and test on each of the five domains (\textit{Books}, \textit{DVD}, \textit{Electronics}, \textit{Kitchen Appliances} and \textit{Movies}) separately. Observing the results, it appears that the effect of $t_{TC}(domain)$ is highest on the domain most correlated with it, suggesting that the adversarial training employed in our Stage 2 intervention, did learn to forget the \textit{TC Topic}. Another interesting pattern is that the estimated effect is higher in more similar domains, and lower on those that are less similar. Specifically, the effect of $t_{TC}(movies)$ is highest on the \textit{Movies} and \textit{DVD} domains, and lowest on the \textit{Electronics} and \textit{Kitchen Appliances} domains. The same pattern can be observed with $t_{TC}(books)$, where the effect is higher on \textit{DVD} and \textit{Movies}, and lower on \textit{Kitchen Appliances} and \textit{Electronics}.

\begin{table}[htbp]
    \centering
    \begin{tabular}{l|l|c|c|c|c|c} 
    \toprule
    Domain & Experiment & $\text{TReATE}_{b}$ & $\text{TReATE}_{d}$ & $\text{TReATE}_{e}$ & $\text{TReATE}_{k}$ & $\text{TReATE}_{m}$ \\
    \midrule
    \multirow{3}{*}{Books} & Balanced & $\textbf{0.131}$ & $0.113$ & $0.034$ & $0.085$ & $0.113$ \\
     & Gentle & $\textbf{0.207}$ & $0.191$ & $0.155$ & $0.156$ & $0.178$ \\
     & Aggressive & $\textbf{0.656}$ & $0.176$ & $0.154$ & $0.137$ & $0.181$ \\ \midrule \midrule
    \multirow{3}{*}{Movies} & Balanced & $0.185$ & $0.185$ & $0.140$ & $0.147$ & $\textbf{0.207}$ \\
     & Gentle & $0.204$ & $0.235$ & $0.204$ & $0.207$ & $\textbf{0.260}$ \\
     & Aggressive & $0.315$ & $0.492$ & $0.272$ & $0.267$ & $\textbf{0.605}$ \\
    \bottomrule
    \end{tabular}  
    \caption{Results for the effect of the $t_{TC}(books)$ (top) and $t_{TC}(movies)$ (bottom) \textit{Topics} on sentiment classification on product and movie reviews. As we do not have access to the ground truth, we compare $\text{TReATE}(O,CF)$ (denoted in the table as $\text{TReATE}_{domain}$) on each domain separately, and denote $b,d,e,k,m$ for each of the domains: \textit{Books}, \textit{DVD}, \textit{Electronics}, \textit{Kitchen Appliances} and \textit{Movies}, respectively. The domain for which the effect of the \textit{treated concept} is the highest for each experiment is highlighted in bold.}
    \label{tab:results-topics}
\end{table}

To test the effect of controlling for multiple concepts, we take advantage of the multi-concepts setting available in the \textit{Topics} dataset. We focus on the \textit{Movies} domain and train \textit{BERT-CF} to forget $t_{TC}(movies)$, while controlling for additional concepts. We perform this experiment with two control concepts, choosing the additional \textit{control concept} in the same manner we chose the first control (i.e., the second control concept is the third most correlated concept with the movies domain). As can be seen in Table \ref{tab:results-multi_topics}, adding control for an additional confounding \textit{Topic} does not substantially affect the $\text{TReATE}_{m}$ estimator. As expected,  $\text{TReATE}_{m}$ does decrease as we add more controls, suggesting that some of the effect captured in the single control experiments is due to confounding information.

\begin{table}[htbp]
    \centering
    \begin{tabular}{l|c|c} 
    \toprule
    & \multicolumn{2}{c}{\# Control Concepts} \\
    Experiment & 1 & 2 \\ \midrule
    Balanced & 0.207 & 0.190\\
    Gentle & 0.260 & 0.247 \\
    Aggressive & 0.605 & 0.607 \\
    \bottomrule
    \end{tabular}  
    \caption{Results for the effect of the $t_{TC}(movies)$ \textit{Topic} on sentiment classification of movie reviews, with one and two \textit{control concepts}.}
    \label{tab:results-multi_topics}
\end{table}

\subsection{Analyzing the Counterfactual Model}
\label{subsec:results-analysis}

Apart from testing the ability of our method to accurately estimate the $\text{ATE}_{gt}(O)$, we want to test the effect of our Stage 2 intervention on the resulting task classifier. Specifically, we look at three aspects, corresponding to questions $2-4$ posed in Section \ref{sec:tasks}. First, we test the accuracy classifiers that utilize the different representation models in predicting the \textit{treated concept} that was adversarially removed in Stage 2 of \textit{BERT-CF}, to check whether we have successfully deleted the concept-related information (question $\#2$). Second, we look at the accuracy of models trying to predict the control concepts, to test that we did not delete information regarding other concepts (question $\#3$). Finally, we look at the performance of models trained in the \textit{Aggressive} setting on a \textit{Balanced} test-set, to test whether we can debias the models using the counterfactual representation and therefore improve the classification accuracy (question $\#4$).

\paragraph{Detecting the Treated Concepts}

\begin{figure}
    \centering
    \includegraphics[width=0.8\columnwidth]{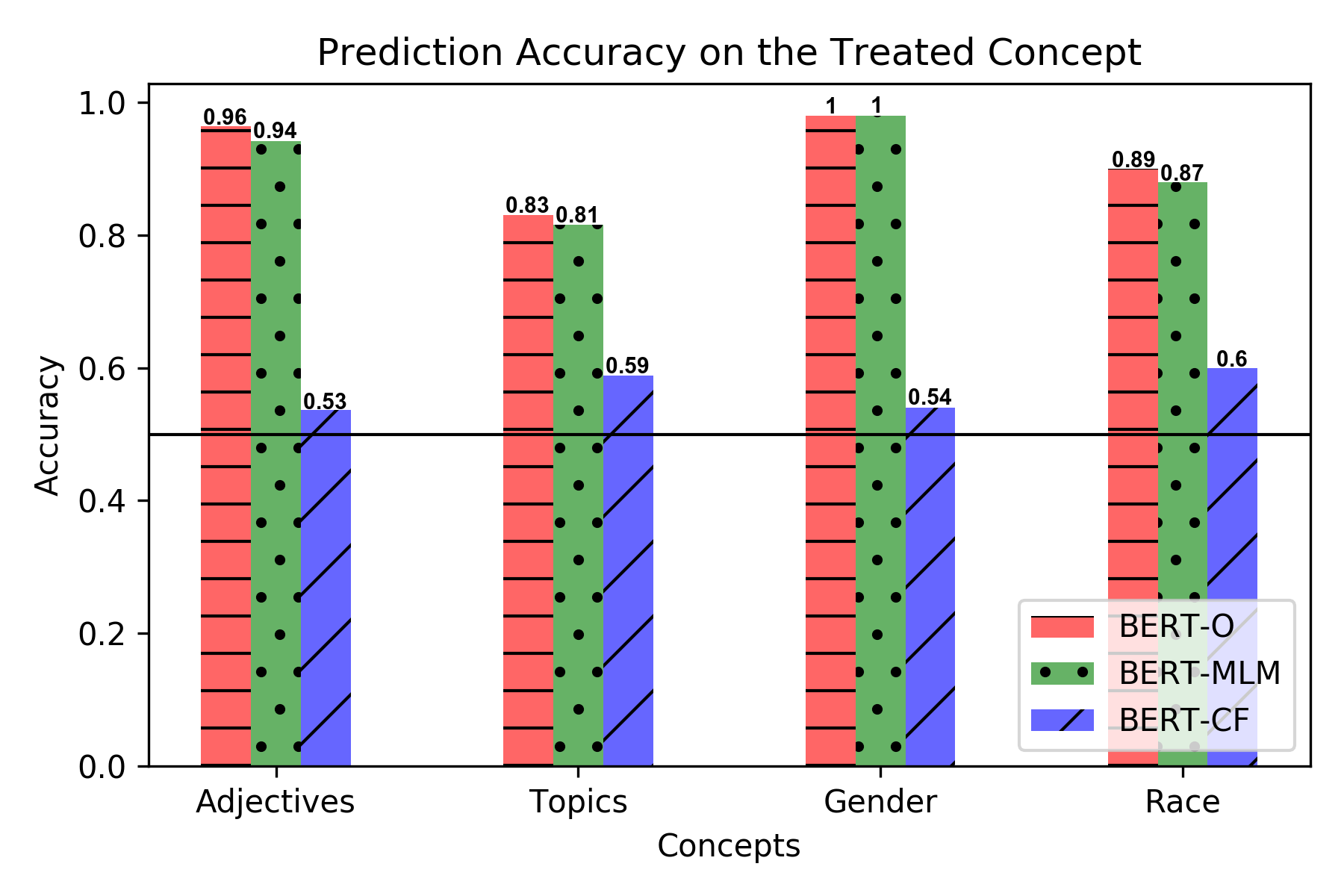}
    \caption{Prediction accuracy on the \textit{treated concept}, averaged over all three dataset versions (\textit{Balanced}, \textit{Gentle} and \textit{Aggressive}), for each concept. For each of the four concepts, we report prediction accuracy on the \textit{treated concept} for classifiers based on \textit{BERT-O}, \textit{BERT-MLM} or \textit{BERT-CF} representations.}
    \label{fig:treated}
\end{figure}

To show that we have successfully trained our \textit{BERT-CF} representation model to forget the \textit{treated concept} (question $\#2$), we compare the accuracy of TC classifiers that use the \textit{BERT-O} representation, the \textit{BERT-MLM} representation, or the treated \textit{BERT-CF} representation. As can be seen in Figure \ref{fig:treated}, the performance of the TC classifiers is very high when using the \textit{BERT-O} or \textit{BERT-MLM} representations, with some achieving almost $100\%$ test-set accuracy. When using the treated representations, however, it is clear that there is a substantial degradation in performance, suggesting that some relevant information was lost. Specifically, in the case of \textit{Gender} and \textit{Adjectives}, and to a lesser extent \textit{Race} and \textit{Topics}, the performance of the \textit{BERT-CF} based TC classifier is only slightly higher than chance.

An important caveat of this analysis is that it does not directly measure the information preserved in the language representation. As discussed in Section \ref{sec:prev}, structural interpretation methods such as those presented here only measure a model's ability to use the representation (in our case, the ability of the TC classifier to use the information encoded in the representation), and not the actual way in which information is encoded in the representation. An analysis of the type presented here should be used as a sanity check, where ground truth data is not available and we want to know whether our counterfactual representation model has forgotten some information about the treated concept. 

\paragraph{Detecting the Control Concepts}

\begin{figure}
    \centering
    \includegraphics[width=0.8\columnwidth]{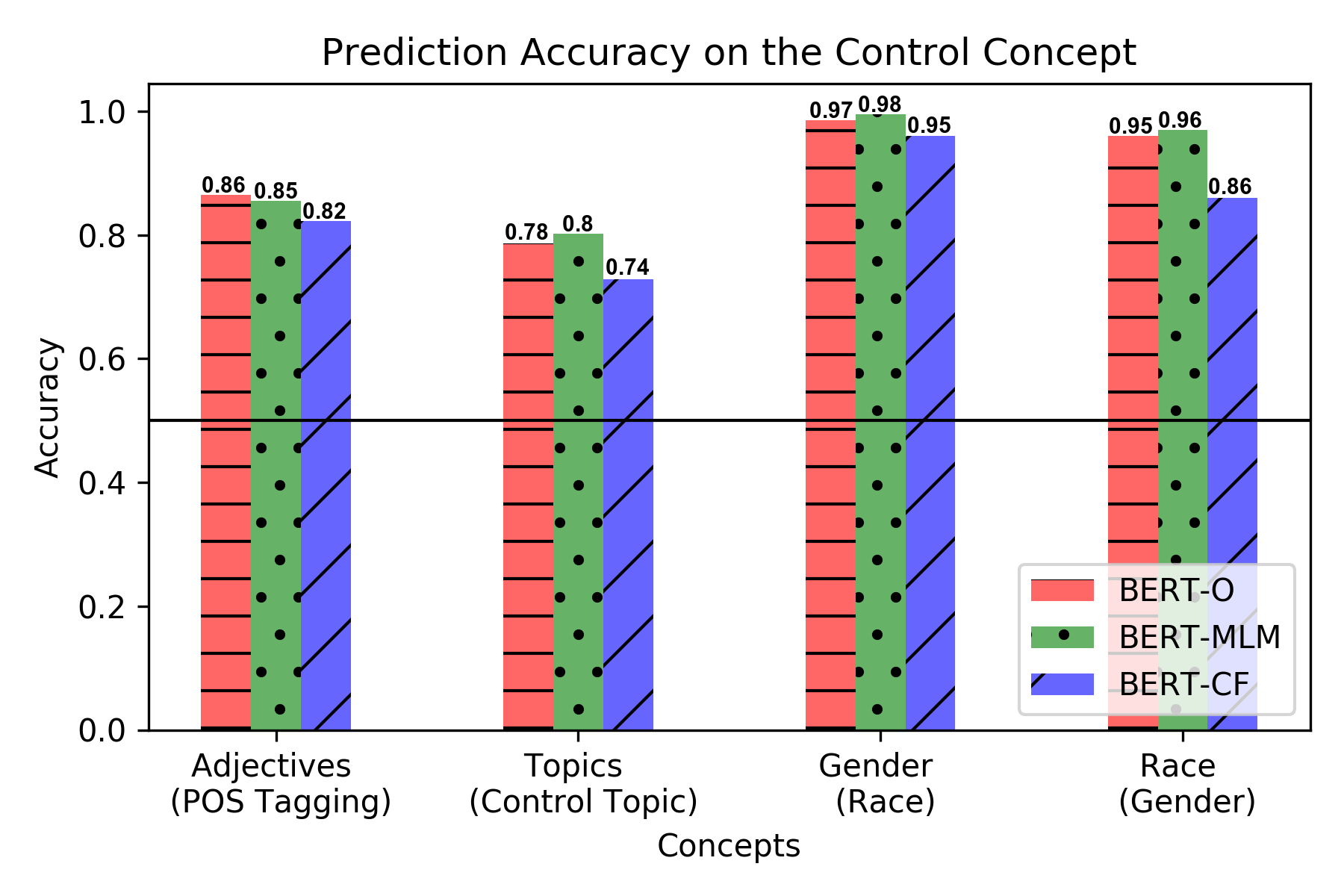}
    \caption{Prediction accuracy on the \textit{control concept}, averaged over all three dataset versions (\textit{Balanced}, \textit{Gentle} and \textit{Aggressive}), for each \textit{control concept}. For each of the four concepts, we report prediction accuracy on the task in parentheses for classifiers based on \textit{BERT-O}, \textit{BERT-MLM} or \textit{BERT-CF} representations.}
    \label{fig:confounders}
\end{figure}

While we have shown that \textit{BERT-CF} has forgotten some information regarding the \textit{treated concept}, it could be at the expense of other concepts correlated with it. So, another useful test for \textit{BERT-CF} is to check if the accuracy of a \textit{BERT-CF} based classifier trained to predict the \textit{control concepts} is the same as the accuracy of corresponding classifiers trained with \textit{BERT-O} or \textit{BERT-MLM} representations. As shown in Figure \ref{fig:confounders}, using the representations from the adversarially trained \textit{BERT-CF} does not hurt performance on related, potentially confounding \textit{control concepts}, showing that we have also successfully answered question $\#3$. In all experimental setups and for all treated concepts, we observe that the difference in performance when using the \textit{BERT-O}, \textit{BERT-MLM} or \textit{BERT-CF} representations is very small. Indeed, using the treated representation degrades performance by only $2-10\%$ (absolute) in terms of accuracy, compared to using the \textit{BERT-O} representation. 

While these results support the claim that these specific confounders were not affected by our Stage 2 intervention, it could very well be that others were affected. This analysis should guide researchers trying to estimate causal effects, and can be used to refute hypotheses regarding specific confounders. In our \textit{Gender} and \textit{Topics} experiments we have created our datasets such that they do not have additional confounders, but for the \textit{Adjectives} and \textit{Topics} experiments there might well be other confounders apart from those tested here.

\paragraph{Mitigating Bias}

The claim in the literature that models pick correlations observed in the training set and are easily biased~\cite{tshitoyan2019unsupervised, gonen2019lipstick} is supported in our experiments. Specifically, looking at the $\text{ATE}_{gt}(O)$ on all experiments with our \textit{Aggressive} setting, it seems that models learned to associate the \textit{treated concept} with the labels. An advantage of our method is that it generates an unbiased language representation with respect to some concept of interest, which can be useful for mitigating such bias.

To test whether we can indeed mitigate that bias (question $\#4$), we test all models trained in the \textit{Aggressive} setting on a \textit{Balanced} test-set. Through these experiments, we can test the generalization of the \textit{BERT-O}, \textit{BERT-MLM} and \textit{BERT-CF} based models. Looking at Table \ref{tab:results-debias}, it is clear that the \textit{BERT-CF} based models can generalize better, and outperform both \textit{BERT-O} and \textit{BERT-MLM} based models when the correlations picked up in training do not exist in the test set. Comparatively, \textit{BERT-CF} is not as affected by the distribution shift, and performs well when the correlation between the \textit{treated concept} and the label changes.

\begin{table}[htbp]
    \centering
    \begin{tabular}{l|c|c|c|c} 
    \toprule
    Concept & Task & \textit{BERT-O} & \textit{BERT-MLM} & \textit{BERT-CF} \\
    \midrule
    Adjectives & Sentiment & $0.75$ & $0.744$ & $\textbf{0.793}$ \\
    Topics & Sentiment & $0.584$ & $0.564$ & $\textbf{0.742}$ \\
    Gender & POMS & $0.924$ & $0.918$ & $\textbf{0.971}$ \\
    Race & POMS & $0.922$ & $0.919$ & $\textbf{0.97}$ \\
    \bottomrule
    \end{tabular}
    \caption{Accuracy of the \textit{BERT-O}, \textit{BERT-MLM} and \textit{BERT-CF} based classifiers when trained in the \textit{Aggressive} settings and tested on the \textit{Balanced} test-set. For each concept the model with the highest accuracy is highlighted in bold.}
    \label{tab:results-debias}
\end{table}

\subsection{Analyzing the Stage 2 Multi-Task Training Scheme}
\label{subsec:losses}
\begin{figure}
    \centering
    \includegraphics[width=0.8\columnwidth]{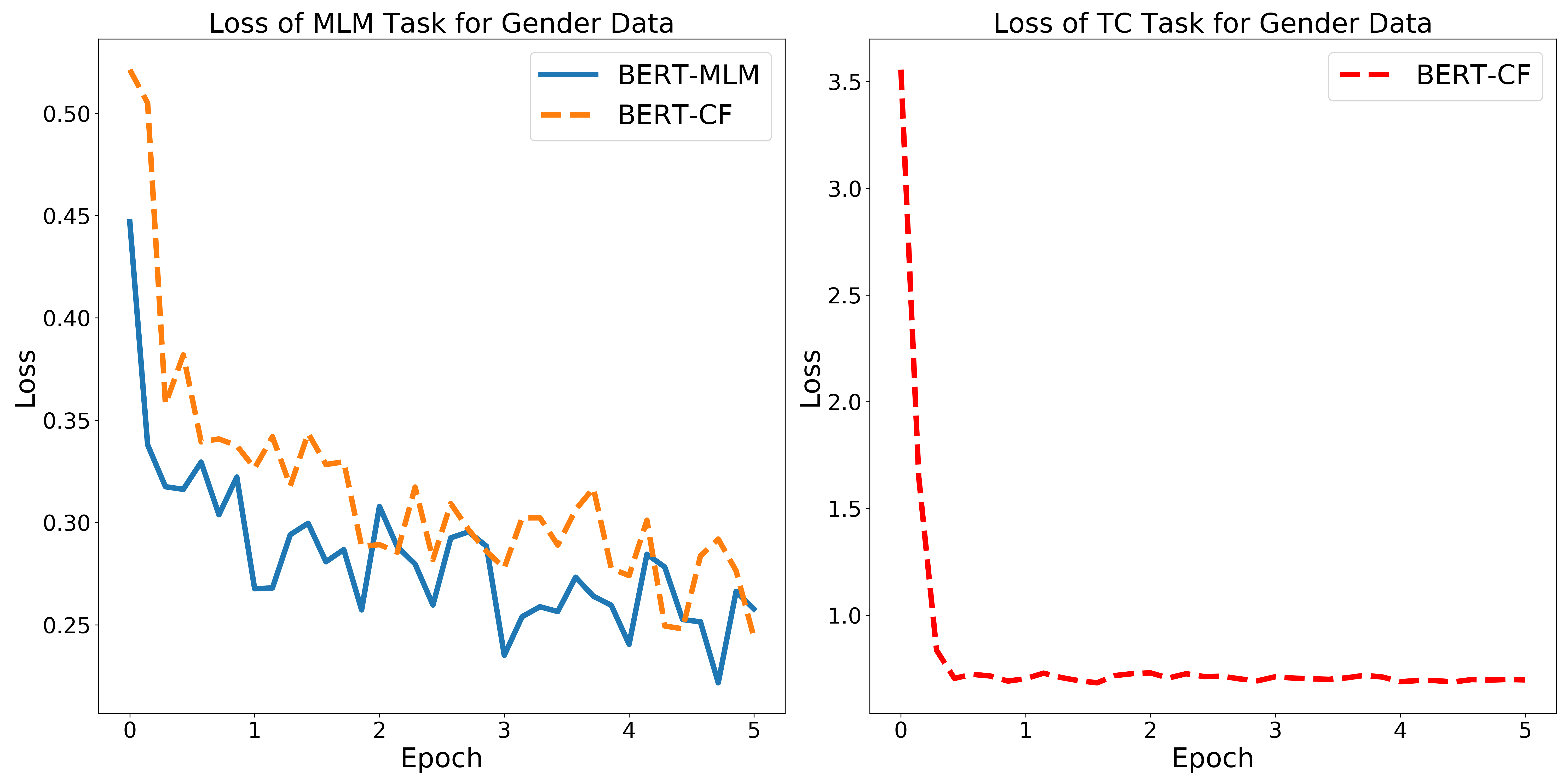}
    \caption{Mean MLM (left) and TC (right) losses (each point is the average of a $1,000$ training steps) for Stage 2 of the \textit{Gender} treatment. \textit{BERT-MLM} refers to the model variant which is trained only on the \textit{MLM} task. \textit{BERT-CF} refers to the model variant which employs both \textit{MLM} and \textit{TC} tasks.}
    \label{fig:gender_mean_loss}
\end{figure}

\begin{figure}
    \centering
    \includegraphics[width=0.8\columnwidth]{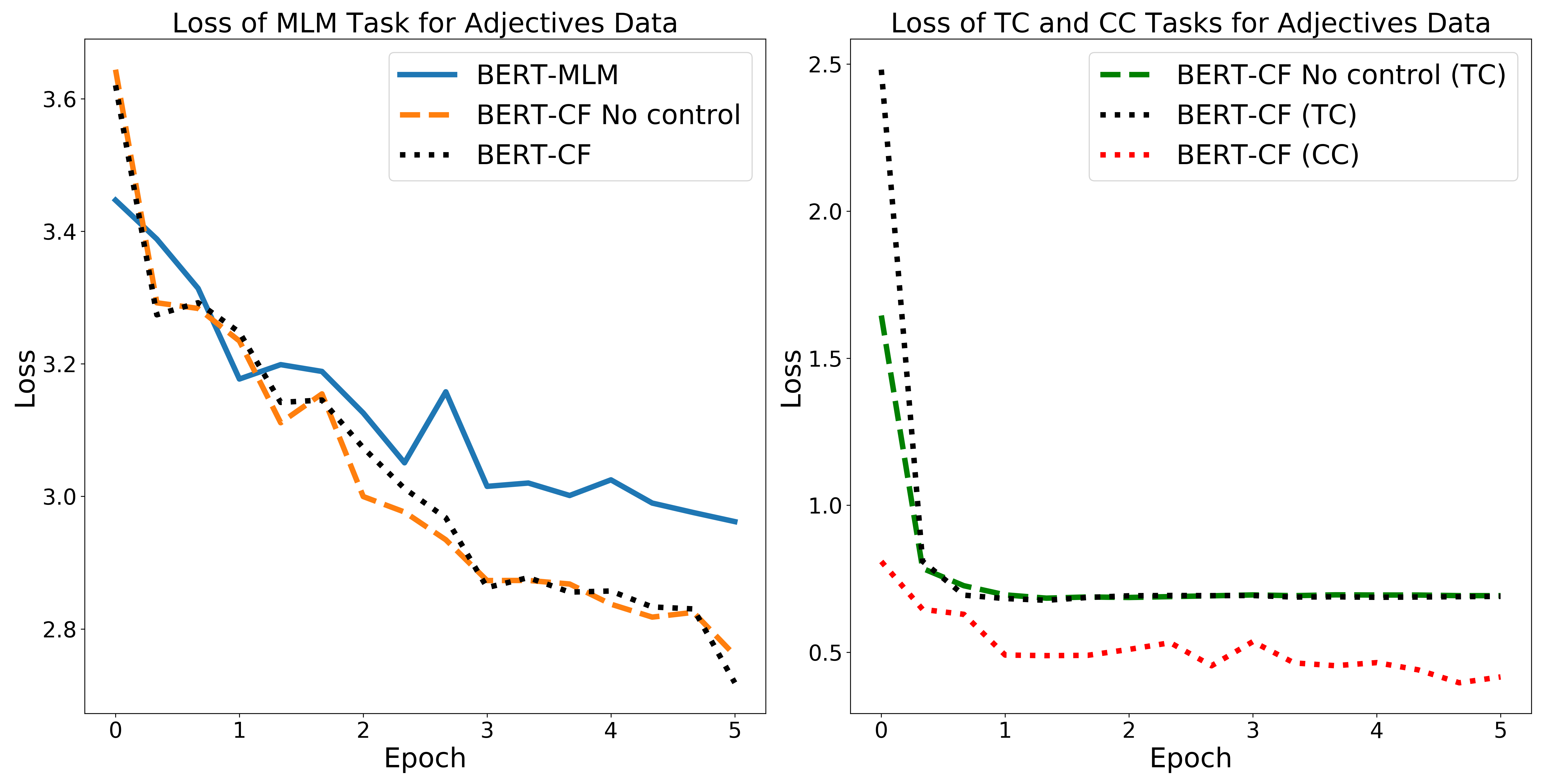}
    \caption{Mean MLM (left) and TC/CC (right) losses (each point is the average of a $1,000$ training steps) for Stage 2 of the \textit{Adjectives} treatment. \textit{BERT-MLM} refers to the model variant which is trained only on the \textit{MLM} task. \textit{BERT-CF} refers to the model variant which employs \textit{MLM} alongside both \textit{TC} and \textit{CC} tasks and  \textit{BERT-CF no control} refers to to the same model without the \textit{CC} task.}
    \label{fig:ima_mean_loss}
\end{figure}

\begin{figure}
    \centering
    \includegraphics[width=0.8\columnwidth]{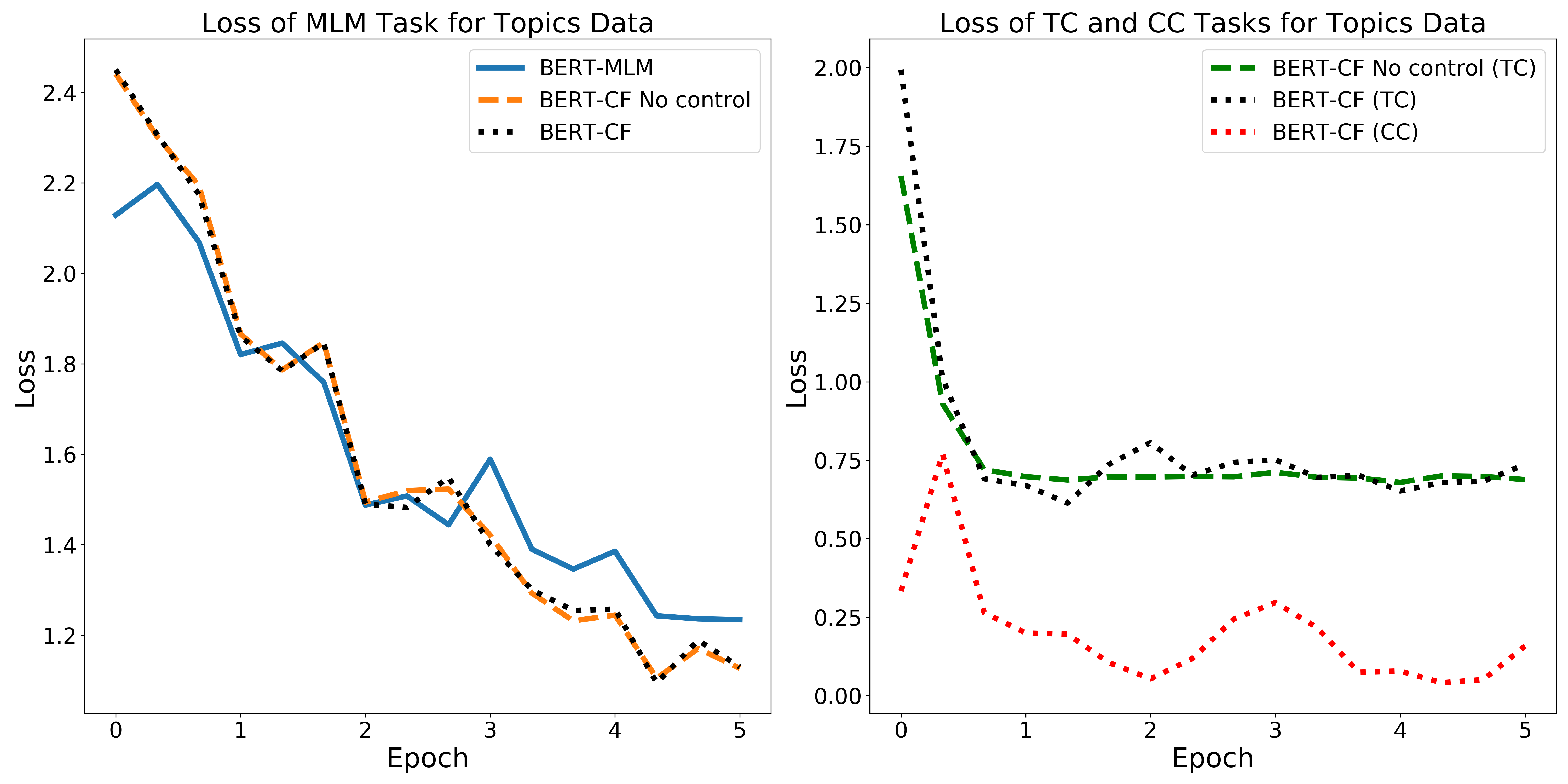}
    \caption{Mean MLM (left) and TC/CC (right) losses (each point is the average of a $1,000$ training steps) for tasks in Stage 2 of the \textit{Topics} treatment. \textit{BERT-MLM} refers to the model variant which is trained only on the \textit{MLM} task.  \textit{BERT-CF} refers to the model variant which employs \textit{MLM} alongside both \textit{TC} and \textit{CC} tasks and  \textit{BERT-CF no control} refers to to the same model without the \textit{CC} task.}
    \label{fig:topics_mean_loss}
\end{figure}

In order to gain further insight into our proposed intervention method, carried out during Stage 2 of the \textit{BERT-CF} training, we present the following loss analysis. We are particularly interested in analyzing the effects of adding \textit{TC} and \textit{CC} tasks to the Stage 2 training scheme, on the \textit{MLM} task and the overall resulting loss function optimization dynamics. While we have shown in Section \ref{subsec:results-analysis} that \textit{BERT-CF} successfully forgets the \textit{TC} task and remembers the \textit{CC} task (questions $\#2$ and $\#3$, respectively), this could be at the expense of the language model. To test if the language model was affected by our Stage 2 intervention, we compare the training losses between \textit{BERT-MLM} and \textit{BERT-CF} without the \textit{CC} task for \textit{Gender} (Figure \ref{fig:gender_mean_loss}), \textit{Adjectives} and \textit{Topics} treatments. For \textit{Adjectives} (Figure \ref{fig:ima_mean_loss}) and \textit{Topics} (Figure \ref{fig:topics_mean_loss}) we also compare to \textit{BERT-CF} with the \textit{CC} task. We do not present figures for \textit{Race} since they are almost identical to the figures for \textit{Gender}.

\begin{figure}
    \centering
    \includegraphics[width=0.8\columnwidth]{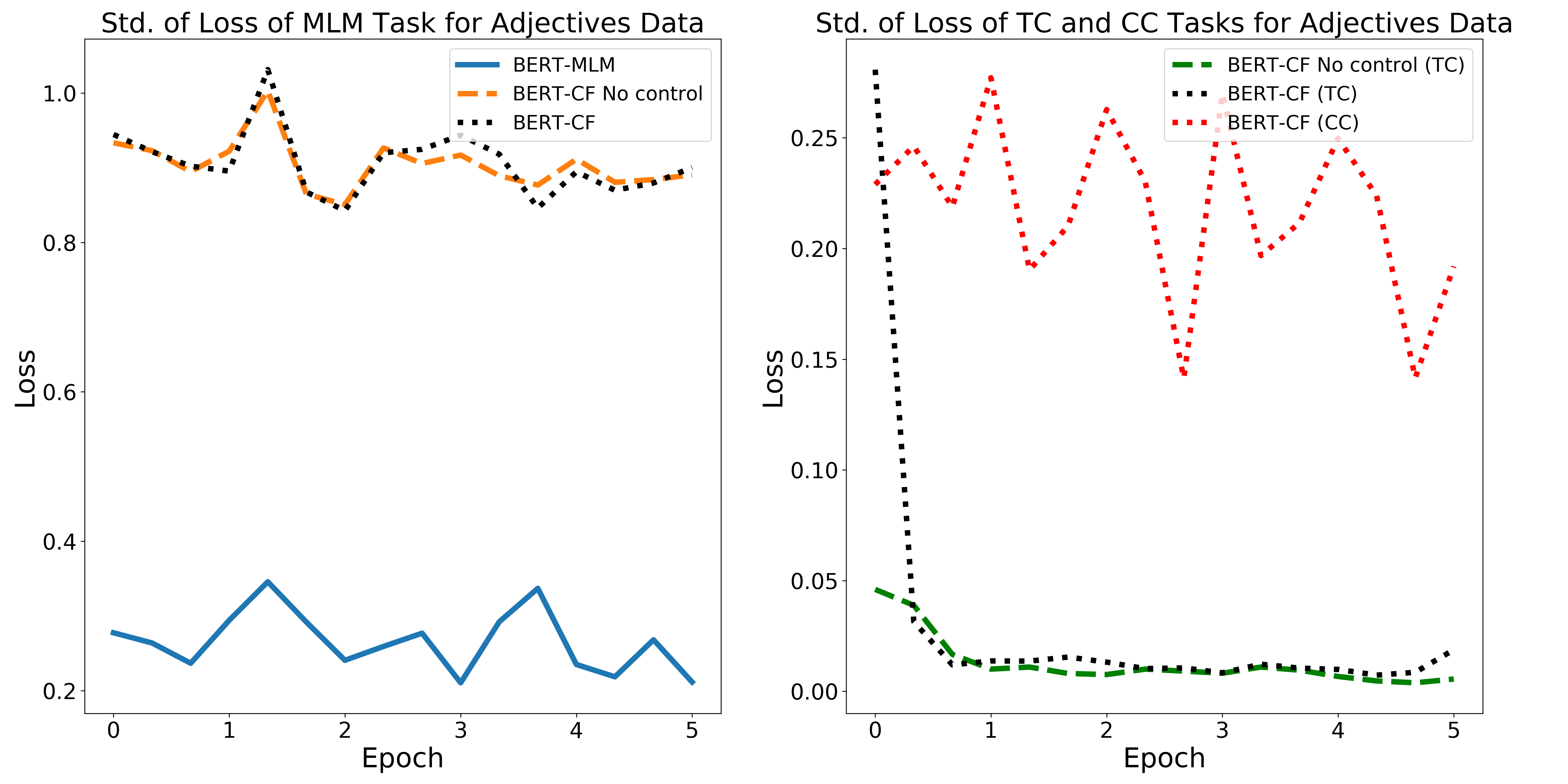}
    \caption{Standard deviation of losses (each point is the average of a $1,000$ training steps) for tasks in Stage 2 of the \textit{Adjectives} treatment. \textit{BERT-MLM} refers to the model variant which is trained only on the \textit{MLM} task. \textit{BERT-CF} refers to the model variant which employs \textit{MLM} alongside both \textit{TC} and \textit{CC} tasks and  \textit{BERT-CF no control} refers to the same model without the \textit{CC} task.}
    \label{fig:ima_std_loss}
\end{figure}

We executed each Stage 2 training process for a total of $5$ epochs, for all variants and all treatments. Our figures present trend lines which are smoothed for visual convenience purposes, by aggregating over batches of $1,000$ training steps rather than over entire epochs. In Figure \ref{fig:ima_std_loss}, we present the standard deviation of the loss values within each $1,000$ training steps only for \textit{Adjectives}, since it best depicts the phenomena occurring for \textit{Gender}, \textit{Race} and \textit{Topics} as well.

The first, most apparent observation we see in all variants and all treatments, is that the \textit{TC} tasks typically converge after $1-2$ epochs. The addition of a \textit{CC} task causes a loss increase during the first epoch of Stage 2 training for all \textit{TC} tasks but quickly converges, typically after $1$ epoch. The \textit{TC} tasks also typically converge to the lowest standard deviation, compared to the \textit{MLM} and \textit{CC} tasks. 

These observations suggest that the addition of a \textit{TC} task introduces an immediate "disturbance" to the BERT encoder, which is expected since the task's goal is to cause the encoder to "forget" features associated with a specific concept. It could be further explained by the dynamics resulting from the adversarial component such \textit{TC} tasks employ, when training alongside "standard" tasks. It is encouraging to see that despite the risk of harming the encoder's representations by adding an adversarial task to the training scheme, this task has an apparent effect without destabilizing all losses which also converge quickly.

For the \textit{Adjectives} treatment, the addition of a \textit{CC} task has a visible effect on the \textit{TC} task, but not on the \textit{MLM} task. The \textit{TC} loss spikes higher at an earlier stage of training, and converges later, in comparison to the \textit{TC} only model variant. Still, the general behavior of the \textit{TC} and \textit{MLM} losses remains very similar in both variants. This suggests that adding the \textit{CC} task dampens the adversarial effect on the \textit{MLM} task, as introduced by the \textit{TC} task, without overwhelming it.

Indeed, it seems that the \textit{CC} task acts as an "opposing" force to the adversarial \textit{TC} task, with the goal of "preserving" features related to a specific concept, while \textit{TC}'s goal is "forgetting" features related to a different concept. We can also see that generally, the \textit{CC} task loss is the quickest to converge in comparison to \textit{TC} and \textit{MLM} tasks. This makes sense, since this task (\textit{PoS Tagging}) is fairly similar to the \textit{TC} task (\textit{IMA}), yet has no adversarial component, and is commonly considered a simpler task compared to \textit{MLM}.

When examining the \textit{MLM} losses across different model variants, we see that all \textit{MLM} mean losses exhibit similar behavior regardless of the variant they come from. While there is an increase in standard deviation in variants which introduce additional tasks, the \textit{MLM} loss is lower in \textit{BERT-CF} compared with \textit{BERT-MLM} in the \textit{Adjectives} and \textit{Topics} experiments, and is only slightly higher in the \textit{Gender} experiments. This suggests that adding well-defined \textit{TC} and \textit{CC} tasks to the Stage 2 training scheme does not drastically harm its stability or the resulting BERT representations.

To summarize, this analysis shows that the addition of the adversarial component in our Stage 2 intervention does not harm the \textit{MLM} or the underlying language representation. Moreover, we have shown that the introduction of the \textit{CC} tasks does not affect the \textit{MLM} as well. While in our analysis in Section \ref{subsec:results-analysis} we have shown that task-trained classifiers using \textit{BERT-CF} can perform well, it could have been due to factors other than the language representation. In the analysis provided here, we have shown that the \textit{MLM} was not harmed by our intervention method, supporting the claim that the language representation remained informative.

%% file: chapters/discussion.tex
\section{Discussion and Conclusion}
\label{sec:discussion}

Our main contributions in this paper are in five directions. First, we have introduced a causal approach for evaluating a variety of hypotheses regarding the effect of a concept on a DNN classifier. Our approach is based on modeling the data-generating process with a causal graph that explicitly states the potential confounding effects between the involved variables. Second, reasoning that direct counterfactual example generation is infeasible with current NLP technology, we have proposed a method for the generation of counterfactual representations, thus avoiding the need for text generation. In causal inference terminology, our method implements the do-operator by adversarially training the language representation. Third, we have created four datasets, each with three variants, where for three of them the true causal effect of a concept can be estimated using manually generated counterfactual examples included in the dataset. Fourth, we have provided tools for evaluating counterfactual language representation models like our \textit{BERT-CF}, in the realistic setup where ground truth causal effect estimations are not available. Finally, we have demonstrated that our counterfactual language representation approach is effective for model debiasing.

Our approach requires making explicit assumptions about the world and generating hypotheses regarding the concepts driving the models' decisions. In order to estimate the impact of a given concept on a DNN, a world model, referred to as a \textit{causal graph} should first be designed. This causal graph depicts the concepts that generate the text that is fed to the DNN, and presents the relations between them. For each concept whose effect we estimated, we have hypothesized how a graph describing the data-generating process might look like, and have approximated its effect relying on this model of the world. In doing so, we have given the modeler a mechanism for explicitly stating her assumptions on the world and the data she is using. While these assumptions are always approximations and are bound to focus on a small number of variables, the alternative is not assumption-free. Indeed, whenever we wish to interpret a model, we are making assumptions on the data-generating process and on the world. Without controlling for confounders, we might end up estimating the effect of variables that are correlated with our \textit{treated concept}. Existing interpretation methods do not explicitly assume a world model, like we do with our causal graphs, but the variables and correlations are still there.

Choosing the \textit{control concepts} is therefore crucial for estimating the true effect and not that of the confounder. Of course, different \textit{control concepts} will probably yield different estimations, and affect the decisions that rely on those interpretations. Using the sanity checks we provide in Section \ref{sec:results}, we have shown that it is possible to test if we have controlled for a given \textit{control concept}. However, it is crucial to identify the treatment and control concepts and form a world model, such as those presented in the causal graphs in Sections \ref{sec:causal} and \ref{sec:tasks}, in order to think through the structure of the task and properly design a CausaLM (or another causal) explanation model.

%

In Appendix B, we present two cases where such world models induce causal graphs with parent concepts that cause other concepts. In such cases, intervening on some concepts will induce a change in concepts that are caused by them, resulting in an estimation that computes both the direct effect of the concept and its effect on concepts which it causes. In areas such as medicine, the causal graph can be built with the assistance of a doctor that is aware of potential confounders and their relationship, but in NLP this is more of a challenge. In Section \ref{subsec:results-analysis} we have proposed several sanity checks that can help modelers understand if their intervention was successful, but in some cases an intervention on a specific concept is not possible.

An important assumption that our world model makes is that concepts can either be switched on or off. Consequently, our representation-based model does not allow us to change the value of a concept (e.g., changing the gender of an example from female to male). Our \textit{TReATE} estimator hence measures the difference between the output class distribution of the classifier in the case where the representation encodes a given concept and the case where the representation does not encode that concept. This, however, is not always in line with the real world. For example, in the \textit{Adjectives} datasets we have created ground truth counterfactual examples where there are no \textit{Adjectives} at all, which is in line with what out \textit{TReATE} estimator measures. However, in cases such as \textit{Gender} and \textit{Race}, for a given example which discusses a person we are interested in the counterfactual where the \textit{Gender} or \textit{Race} of the person is switched (e.g., from male to female) while deleting the concept is not feasible (e.g., gender is encoded in English sentences through words like \textit{him, her, he} and \textit{she}). Our results in Section \ref{sec:results} suggest that we can still estimate the causal effect of the \textit{Gender} and \textit{Race} concepts on the classifier with \textit{TReATE}, although it compares the output class distribution of the model between representations that encode information about the concept and representations that omit this information. Despite the empirical success, this is a limitation of our framework that we plan to address in future work.

The discussion above emphasizes the importance of world knowledge and assumptions for interpreting DNNs. As long as assumptions have to be made on the connection between concepts or features, it is crucial that modelers make these assumptions explicitly, and not just implicitly. While these assumptions might be wrong, they can ground the discussion and allow for empirical tests to be made. In the experiments we conducted, we have tried to address many different types of assumptions on what is driving the data-generating process.

Another issue we have discussed is the distinction between global and local concepts. Language expresses many global concepts such as the topics being discussed and the style being used, but these are very hard to model. We proposed an elegant solution for \textit{Topics} that is both global (shares information between the training set sentences) and can be integrated into our counterfactual representation model. Unlike local concepts such as \textit{Gender} and \textit{Race} which are widely researched, there is very little work on understanding the effect of global concepts on DNNs. As they are hard to measure and cannot be computed through a token-level analysis, they remain an open challenge. Understanding the effect of global concepts is a direction that we wish to further explore in future work.

Finally, there is a significant challenge in validating the quality of a causal explanation method. One method, which we have used in our \textit{Gender} and \textit{Race} experiments, is to use synthetic data, where the validation is more accurate and reliable. However, this comes at the expense of not using real-world data, which is more natural and complex. When using real-world data to validate causal methods, we often need to generate counterfactual examples manually. While in the \textit{Adjectives} experiments we were able to create such examples without manual interventions, it is almost always hard to do. For example, it would be almost impossible to create a counterfactual example with respect to a \textit{Topic} without manually generating a new example.

In this work we have created four datasets, two synthetic and two real-world, that allow for such causal validation. However, the synthetic dataset (EEEC), is limited in terms of the language it expresses, and the real-world dataset (Sentiment) only has automatically generated counterfactual examples, which can be inaccurate. We see the creation of a dataset that is both natural and includes precise counterfactual examples as crucial for the advancement and dissemination of causal model explanations in NLP, and plan to further develop such datasets.

%% file: chapters/appendix.tex
\section{A Clinical Example}
\label{app:clinical}

While failing to estimate the causal effect of a concept on a sentiment classifier is harmful, it pales in comparison to the potential harm of wrongfully interpreting clinical prediction models. If a model is trained on clinical notes to predict clinically important factors, the need for understanding the model is amplified. If it were the case that the model is relying on textual features that are doctor or hospital specific, it could lead to devastating implications. 

For example, we can look at the (fake) clinical note presented in Figure \ref{fig:example-clinical}. In this note, the patient's mental health is discussed extensively, with a verbose description and much detail. As the description is lengthy and sometimes repetitive, it could be summarized without losing too much clinically relevant information. In this case, a classifier that heavily relies on the doctor's verbose style could fail when given a short and concise note which still contains the same clinical information.

\begin{figure}[htbp]
    \centering
    \framebox{%
    \begin{varwidth}{0.90\textwidth}
    \underline{Status of patient:}
     
     Julie  \textcolor{green}{\textbf{is}} worse today.
     
     \underline{Target Symptoms:}
     
     Julie reports \textcolor{green}{\textbf{that}} \textcolor{orange}{\textbf{depressive}} symptoms \textcolor{orange}{\textbf{continue.}} \textcolor{green}{\textbf{Her}} \textcolor{orange}{\textbf{symptoms,}} \textcolor{green}{\textbf{she}} reports, \textcolor{green}{\textbf{are}} \textcolor{green}{\textbf{more}} frequent \textcolor{green}{\textbf{or}} \textcolor{green}{\textbf{more}} intense. Anergia \textcolor{green}{\textbf{is}} present. \textcolor{orange}{\textbf{Increased}} symptoms \textcolor{green}{\textbf{of}} \textcolor{orange}{\textbf{anhedonia}} \textcolor{green}{\textbf{are}} present. Julie's \textcolor{orange}{\textbf{difficulty}} \textcolor{green}{\textbf{with}} \textcolor{red}{\textbf{concentrating}} \textcolor{green}{\textbf{has}} \textcolor{green}{\textbf{not}} changed. Julie reports \textcolor{green}{\textbf{that}} \textcolor{green}{\textbf{she}} \textcolor{orange}{\textbf{continues}} \textcolor{green}{\textbf{to}} \textcolor{green}{\textbf{feel}} \textcolor{green}{\textbf{sad.}} Guilty feelings \textcolor{green}{\textbf{are}} \textcolor{orange}{\textbf{described}} \textcolor{green}{\textbf{by}} Julie. \textcolor{green}{\textbf{'I}} should \textcolor{green}{\textbf{have}} \textcolor{green}{\textbf{been}} \textcolor{green}{\textbf{with}} \textcolor{green}{\textbf{my}} sister, \textcolor{green}{\textbf{I}} \textcolor{green}{\textbf{had}} \textcolor{green}{\textbf{no}} \textcolor{green}{\textbf{idea}} \textcolor{green}{\textbf{she}} \textcolor{green}{\textbf{was}} \textcolor{orange}{\textbf{suicidal.'}} Sleep \textcolor{green}{\textbf{has}} improved \textcolor{green}{\textbf{with}} \textcolor{green}{\textbf{the}} \textcolor{green}{\textbf{use}} \textcolor{green}{\textbf{of}} \textcolor{green}{\textbf{PRN}} Ambien \textcolor{green}{\textbf{CR}} \textcolor{green}{\textbf{at}} \textcolor{green}{\textbf{HS.}} Julie \textcolor{red}{\textbf{convincingly}} denies suicidal ideas \textcolor{green}{\textbf{or}} \textcolor{red}{\textbf{intentions.}} 
     
     \underline{Basic \textcolor{orange}{\textbf{Behaviors:}}}
     
     \textcolor{orange}{\textbf{Medication}} \textcolor{green}{\textbf{has}} \textcolor{green}{\textbf{been}} taken \textcolor{orange}{\textbf{regularly.}} \textcolor{green}{\textbf{She}} needs \textcolor{green}{\textbf{help}} \textcolor{green}{\textbf{with}} ADLs. \textcolor{green}{\textbf{When}} \textcolor{green}{\textbf{she}} attends \textcolor{orange}{\textbf{activities}} \textcolor{red}{\textbf{participation}} \textcolor{green}{\textbf{is}} minimal. Prn's \textcolor{green}{\textbf{are}} \textcolor{green}{\textbf{used}} \textcolor{red}{\textbf{occasionally}} \textcolor{green}{\textbf{and}} \textcolor{green}{\textbf{are}} \textcolor{orange}{\textbf{described}} \textcolor{green}{\textbf{as}} \textcolor{orange}{\textbf{effective}} \textcolor{green}{\textbf{for}} \textcolor{green}{\textbf{her}} \textcolor{orange}{\textbf{headaches.}} \textcolor{orange}{\textbf{Impulsive}} \textcolor{orange}{\textbf{behaviors}} \textcolor{green}{\textbf{are}} \textcolor{orange}{\textbf{occurring,}} \textcolor{green}{\textbf{but}} \textcolor{green}{\textbf{less}} \textcolor{red}{\textbf{frequently.}} Julie \textcolor{green}{\textbf{has}} \textcolor{orange}{\textbf{diminished}} \textcolor{green}{\textbf{food}} \textcolor{green}{\textbf{and}} fluid intake. Julie \textcolor{green}{\textbf{has}} \textcolor{green}{\textbf{not}} \textcolor{green}{\textbf{been}} \textcolor{orange}{\textbf{confused.}} \textcolor{green}{\textbf{A}} \textcolor{green}{\textbf{good}} night's sleep \textcolor{green}{\textbf{is}} \textcolor{red}{\textbf{described.}}
     
      \underline{\textcolor{red}{\textbf{Additional}} Signs \textcolor{green}{\textbf{or}} Possible \textcolor{green}{\textbf{Side}} Effects:}
     
     Sedative effects \textcolor{green}{\textbf{of}} \textcolor{green}{\textbf{the}} \textcolor{orange}{\textbf{medication}} \textcolor{green}{\textbf{are}} \textcolor{orange}{\textbf{described.}} Patient reports \textcolor{green}{\textbf{a}} \textcolor{green}{\textbf{dry}} mouth. \textcolor{green}{\textbf{No}} other \textcolor{green}{\textbf{side}} effects \textcolor{green}{\textbf{are}} reported \textcolor{green}{\textbf{or}} \textcolor{green}{\textbf{in}} \textcolor{orange}{\textbf{evidence.}} 
     
     \underline{MENTAL STATUS:}
     
     Julie presents \textcolor{green}{\textbf{as}} glum, \textcolor{orange}{\textbf{downcast,}} \textcolor{red}{\textbf{inattentive,}} \textcolor{orange}{\textbf{minimally}} \textcolor{red}{\textbf{communicative,}} \textcolor{green}{\textbf{and}} looks unhappy. \textcolor{green}{\textbf{She}} appears listless \textcolor{green}{\textbf{and}} anergic. \textcolor{green}{\textbf{She}} appears \textcolor{orange}{\textbf{downcast.}} Thought content \textcolor{green}{\textbf{is}} \textcolor{orange}{\textbf{depressed.}} Slowness \textcolor{green}{\textbf{of}} physical movement helps reveal \textcolor{orange}{\textbf{depressed}} mood. Facial \textcolor{orange}{\textbf{expression}} \textcolor{green}{\textbf{and}} general demeanor reveal \textcolor{orange}{\textbf{depressed}} mood. \textcolor{green}{\textbf{She}} denies having suicidal ideas. There \textcolor{green}{\textbf{are}} \textcolor{green}{\textbf{no}} apparent signs \textcolor{green}{\textbf{of}} \textcolor{red}{\textbf{hallucinations,}} \textcolor{orange}{\textbf{delusions,}} bizarre \textcolor{orange}{\textbf{behaviors,}} \textcolor{green}{\textbf{or}} other \textcolor{orange}{\textbf{indicators}} \textcolor{green}{\textbf{of}} \textcolor{orange}{\textbf{psychotic}} process. \textcolor{red}{\textbf{Associations}} \textcolor{green}{\textbf{are}} intact, thinking \textcolor{green}{\textbf{is}} logical, \textcolor{green}{\textbf{and}} thought content appears \textcolor{red}{\textbf{appropriate.}} There \textcolor{green}{\textbf{are}} signs \textcolor{green}{\textbf{of}} anxiety. Patient \textcolor{green}{\textbf{is}} fidgety \textcolor{green}{\textbf{in}} \textcolor{green}{\textbf{a}} \textcolor{green}{\textbf{way}} \textcolor{green}{\textbf{that}} \textcolor{green}{\textbf{is}} \textcolor{orange}{\textbf{suggestive}} \textcolor{green}{\textbf{of}} \textcolor{red}{\textbf{anxiety.}} 
     
     \underline{Special \textcolor{red}{\textbf{Circumstances:}}}
     
     Julie \textcolor{orange}{\textbf{continues}} \textcolor{green}{\textbf{to}} \textcolor{green}{\textbf{have}} \textcolor{green}{\textbf{an}} unsteady gait, \textcolor{orange}{\textbf{especially}} after \textcolor{orange}{\textbf{midnight.}} \textcolor{green}{\textbf{Call}} light \textcolor{green}{\textbf{is}} within \textcolor{green}{\textbf{her}} reach. \textcolor{green}{\textbf{She}} \textcolor{green}{\textbf{has}} \textcolor{green}{\textbf{been}} \textcolor{orange}{\textbf{instructed}} \textcolor{green}{\textbf{to}} \textcolor{green}{\textbf{ring}} \textcolor{green}{\textbf{for}} \textcolor{green}{\textbf{the}} nurse \textcolor{green}{\textbf{to}} assist \textcolor{green}{\textbf{her}} \textcolor{green}{\textbf{when}} \textcolor{orange}{\textbf{ambulating}} \textcolor{green}{\textbf{to}} \textcolor{red}{\textbf{bathroom.}} 
     
     \underline{Vital Signs:}
     
     Sitting blood pressure \textcolor{green}{\textbf{is}} \textcolor{green}{\textbf{150}} \\ \textcolor{green}{\textbf{/}} \textcolor{green}{\textbf{85.}} Sitting pulse \textcolor{green}{\textbf{rate}} \textcolor{green}{\textbf{is}} \textcolor{green}{\textbf{80.}} \textcolor{red}{\textbf{Respiratory}} \textcolor{green}{\textbf{rate}} \textcolor{green}{\textbf{is}} \textcolor{green}{\textbf{18}} \textcolor{green}{\textbf{per}} minute. Temp. \textcolor{green}{\textbf{is}} \textcolor{green}{\textbf{98+}} \textcolor{green}{\textbf{F.}} Weight \textcolor{green}{\textbf{is}} \textcolor{green}{\textbf{155}} \textcolor{green}{\textbf{lbs.}} (70.3 \textcolor{green}{\textbf{Kg}}).
    \end{varwidth}%
    }
    \caption{A fake example of a \textit{Nursing Progress Note} taken from \url{https://www.examples.com/business/progress-note.html}. Highlighted in \textcolor{red}{\textbf{red}} and \textcolor{orange}{\textbf{orange}} are words with length in the $90-100$ and $75-90th$ quantiles, respectively. \textcolor{green}{\textbf{Green}} words are of length that is below the $25th$ quantile. Qunatiles are measured based on the frequency of all words in the clinical note.}
    \label{fig:example-clinical}    
\end{figure}

In this note, we highlight words by their length, a feature described previously as a proxy for writing style~\cite{sari2018topic}. Looking at the words highlighted in red and in orange, it is clear that changing the writing style of the note would require a significant intervention. At the same time, deleting long words or replacing them would significantly affect the structure and content of the note. Moreover, different note's sections tend to have words with different lengths (for example, the \textit{Vital Signs} section contains very short words), which could be a potential confounder. If we intervene and replace longer words with shorter synonyms, we might also change some concepts alongside the ones we mean to change, and there is no test that will tell us that ex-ante. 

Concepts that influence both the label and other concepts, also known as \textit{confounders}, could be extremely risky. Imagine a case where a doctor receives on average more patients that are of certain type, such as individuals with severe depression. In that case, a DNN could learn to associate this writing style as a signal for a depressed patient. Measuring the model's performance on notes written by that doctor would show promising results, but deploying such a model would risk patients health when used on patients of a different doctor. Without measuring the causal effect of the doctor's writing style on the classifier, we would not be able to tell to what extent the model is relying on it.~\footnote{Note that while clinical notes are an important application domain, we do not consider them in our experiments as they were not publicly available to us. We plan to create such synthetic data in future work.}

Other, more complex relationships, might exist between concepts. For example, if a clinical note is describing the doctor's clinical treatment suggestion based on the patient's condition (i.e., depression, anxiety etc.), it would be hard to disentangle the clinical treatment suggestion from a specific condition (the causal graph for this example is presented in the bottom graph in Figure \ref{fig:cg-clinical_note}). Alternatively, it could also be that only the patient's depression or lack of it is causing the doctor's treatment suggestion, and the text is generated based on that suggestion alone (see the bottom graph in Figure \ref{fig:cg-clinical_note}). This would make it impossible to imagine a counterfactual text, where the upstream concept (depression) is changed but the one generated by it (treatment suggestion) remains fixed. We next (Appendix B)  discuss alternative causal graphs that can be modeled, and highlight the power and limitations of using a world model such as those discussed here to interpret DNNs.

\section{Alternative Causal Graphs and Limitations}
\label{app:alt-graph}

\begin{figure}[!htbp]
 \centering
 \includegraphics[width=0.6\linewidth]{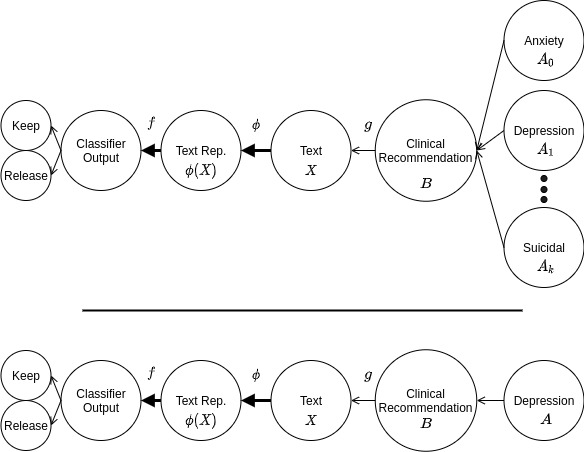}    
    \caption{Two plausible causal graphs for a case where a patient's condition (\textit{Anxiety}, \textit{Depression} and \textit{Suicidal}) is causing the doctor's \textit{Clinical Recommendation}, which is then generating a text. The classifier then uses this text to decide if the patient should be kept in the hospital (\textit{Keep}) or released (\textit{Release}). The top graph represents a data-generating process where many conditions cause the doctor's \textit{Clinical Recommendation}, such as \textit{Anxiety}, \textit{Depression} and \textit{Suicidal}. The bottom graph represents the scenario where only \textit{Depression} causes the \textit{Clinical Recommendation}.}
    \label{fig:cg-clinical_note}    
\end{figure}

Our ability to intervene on the \textit{treated concept} and estimate $\widehat{TReATE}_{TC}$ is dependent on the world model we assume, as presented in the causal graph. For the examples presented in Figures \ref{fig:example} and \ref{fig:example-clinical} we have suggested causal graphs (Figure \ref{fig:cg-example}) where the relationship between the concepts generating the text is rather simple, as all concepts generate the text without any inheritance relations (i.e where one concept causes the other). In many interesting cases, the relationship between concepts is not as straight-forward, and might affect our ability to intervene on some concepts. In Figure \ref{fig:cg-clinical_note}, we consider two such cases, where one or more concepts $A_j | j \in \{0,1, \ldots, k \}$ cause a concept $B$, which in turn generates a text $X$. In the clinical note example, it could be the case that a patient's condition is causing a doctor to recommend a specific treatment, and this decision induces the doctor to write the note. In Figure \ref{fig:cg-clinical_note} we propose two causal graphs that model this data-generating process, for the case where many conditions cause the doctor's decision (top graph) and for the case where only the patient's depression affects the decision (bottom graph).

Intervening on specific concepts and computing the causal effect of those concepts on the decisions made by the classifier is not as straight-forward. In the first case, where \textit{Anxiety}, \textit{Depression} and \textit{Suicidal} are causing the doctor's \textit{Clinical Recommendation}, intervening on \textit{Depression} would also affect the \textit{Clinical Recommendation}, as it is one of its causes. While our method can accommodate for this case, our estimator will measure both the direct effect of \textit{Depression} and its indirect effect, through the \textit{Clinical Recommendation}. In such a case it would be impossible to preserve all the information resulting from the \textit{Clinical Recommendation} concept while omitting the information from the \textit{Depression} concept, due to the inheritance relation between the two concepts. Intervening on the \textit{Clinical Recommendation} will be even more problematic, as a text without a recommendation will have to be blank according to this graph, regardless of the patient's underlying condition.

The causal graph presented in the bottom of Figure \ref{fig:cg-clinical_note} is also possible, meaning that it could be a reasonable world model in some cases. However, if we were to intervene on \textit{Depression}, we will not be able to know what the doctor would have recommended. In this case we could not estimate the causal effect of \textit{Depression} or that of the \textit{Clinical Recommendation} on the classifier's decision. In Section \ref{subsec:results-analysis} we suggest several sanity checks to help modelers understand if the world model they are using has successfully learned to forget the \textit{treated concept} while remembering the \textit{control concepts}, but there are certainly cases where the ability to perform such interventions is limited.

\section {The EEEC Dataset}
\label{data-tables}

As discussed in Section \ref{subsec:eeec} we provide here additional details about the generation of the EEEC dataset. Particularly, in Table \ref{tab:eeec-templates} we present the templates used to generate the data, and in Table \ref{tab:eeec-data} we compare the original EEC and our EEEC, to illustrate the key modifications we have made.

\begin{table}
    \centering
    \small
    \begin{tabular}{|p{0.3cm}|p{11cm}|p{1cm}|} \hline
        ID & Template & \# Sent. \\ \hline
        1 & Now that it is all over, <person> feels <emotion> & $787$ \\
        2 & <person> feels <emotion> as <gender noun> walks to the <place> & $490$ \\
        3 & Even though it is still a work in progress, the situation makes <person> feel <emotion> & $286$ \\
        4 & The situation makes <person> feel <emotion>, and will probably continue to in the foreseeable future & $1,145$ \\
        5 & It is a mystery to me, but it seems i made <person> feel <emotion> & $598$ \\
        6 & I made <person> feel <emotion>, and plan to continue until the <season> is over & $1,114$ \\
        7 & It was totally unexpected, but <person> made me feel <emotion> & $691$ \\
        8 & <person> made me feel <emotion> for the first time ever in my life & $1,218$ \\
        9 & As <gender noun> approaches the <place>, <person> feels <emotion> & $1,504$ \\
        10 & <person> feels <emotion> at the end & $598$ \\
        11 & While it is still under construction, the situation makes <person> feel <emotion> & $400$ \\
        12 & It is far from over, but so far i made <person> feel <emotion> & $531$ \\
        13 & We went to the <place>, and <person> made me feel <emotion> & $891$ \\
        14 & <person> feels <emotion> as <gender noun> paces along to the <place> & $550$ \\
        15 & While this is still under construction, the situation makes <person> feel <emotion> & $335$ \\
        16 & The situation makes <person> feel <emotion>, but it does not matter now & $1,131$ \\
        17 & There is still a long way to go, but the situation makes <person> feel <emotion> & $312$ \\
        18 & I made <person> feel <emotion>, time and time again & $1,188$ \\
        19 & While it is still under development, the situation makes <person> feel <emotion> & $261$ \\
        20 & I do not know why, but i made <person> feel <emotion> & $492$ \\
        21 & <person> made me feel <emotion> whenever I came near & $1,092$ \\
        22 & While we were at the <place>, <person> made me feel <emotion> & $648$ \\
        23 & <person> feels <emotion> at the start & $483$ \\
        24 & Even though it is still under development, the situation makes <person> feel <emotion> & $285$ \\
        25 & I have no idea how or why, but i made <person> feel <emotion> & $468$ \\
        26 & We were told that <person> found <gender noun> in <ind> <emotion> situation & $1,168$ \\
        27 & <person> found <gender noun> in <ind> <emotion> situation, after <time> & $1,164$ \\
        28 & As we were walking together, <person> told us all about the recent <emotion> events & $1,164$ \\
        29 & <person> told us all about the recent <emotion> events as we were walking to the <place> & $1,156$ \\
        30 & As expected, the conversation with <person> was <emotion> & $728$ \\
        31 & The conversation with <person> was <emotion>, we could from simply looking & $1,128$ \\
        32 & To our surprise, <person> found <gender noun> in <ind> <emotion> situation & $1,152$ \\
        33 & <person> found <gender noun> in <ind> <emotion> situation, something none of us expected & $1,156$ \\
        34 & While we were walking to the <place>, <person> told us all about the recent <emotion> events & $1,156$ \\
        35 & The conversation with <person> was <emotion>, you could feel it in the air & $1,192$ \\
        36 & While unsurprising, the conversation with <person> was <emotion> & $748$ \\
        37 & <person> told us all about the recent <emotion> events, to our surprise & $1,164$ \\
        38 & To our amazement, the conversation with <person> was <emotion> & $844$ \\
        39 & I <observe> <person> in the <place> <day>. & $580$ \\
        40 & I talked to <person> <day>. & $580$ \\
        41 & <person> goes to the school in our neighborhood. & $580$ \\
        42 & <person> has <number> <family>. & $580$ \\ \hline
    \end{tabular}
    \caption{The templates used to generate the EEEC dataset.}
    \label{tab:eeec-templates}
\end{table}

\begin{table}[!h]
    \centering
    \begin{tabular}{|l|c|c|} \hline
        Metric & EEC & EEEC \\ \hline
        Full Data Size (\# of Sentences) & $9,840$ & $33,738$ \\
        Median Sentence Length (\# of words) & $6$ & $14$ \\
        \# of Templates & $11$ & $42$ \\
        \# of Noise Sentences & $0$ & $13$ \\
        \# of Prefix Sentences & $0$ & $21$ \\
        \# of Suffix Sentences & $0$ & $16$ \\
        \# of Emotion Words & $40$ & $55$ \\
        \# of Female Names & $10$ & $10$ \\
        \# of Male Names & $10$ & $10$ \\
        \# of European Names & $10$ & $10$ \\
        \# of African-American Names & $10$ & $10$ \\
        \# of Places & $10$ & $10$ \\ \hline
    \end{tabular}
    \caption{Descriptive statistics comparing the EEC and Enriched EEC (EEEC) datasets.}
    \label{tab:eeec-data}
\end{table}

\begin{table}[!h]
    \centering
    \begin{tabular}{|l|c|} \hline
       Hyper-parameter & \# \\ \hline
       Random Seed & $212$ \\ \hline
       Sentiment maximum sequence length & $384$ \\ \hline
       POMS maximum sequence length & $32$ \\ \hline
       Stage 2 \textit{TC} (adversarial) task $\lambda$ & $1$ \\ \hline
       Stage 2 number of epochs & $5$ \\ \hline
       Stage 2 Sentiment batch size & $6$ \\ \hline
       Stage 2 POMS batch size & $24$ \\ \hline
       Stage 3 number of epochs & $50$ \\ \hline
       Stage 3 Sentiment batch size & $128$ \\ \hline
       Stage 3 POMS batch size & $200$ \\ \hline
       Stage 3 gradient accumulation steps & $4$ \\ \hline
       Stage 3 classifier dropout probability & $0.1$ \\ \hline
    \end{tabular}
    \captionsetup{justification=centering}
    \caption{The hyper-parameters used in our experiments.}
    \label{tab:bert_params}
\end{table}

\section{Experimental Pipeline and Hyper-parameters
\label{app:params}}

The pipeline of our experiments follows the same steps for all the settings we address. Particularly, we execute the following pipeline for each of the downstream classification tasks (\textit{Sentiment} (Section \ref{sec:adjectives} and \ref{sec:topics}) and \textit{POMS} (Section \ref{sec:poms}), as well as for the \textit{TC} and \textit{CC} probing tasks (Section \ref{subsec:results-analysis})) and for each version of the datasets (\textit{Balanced}, \textit{Gentle} and \textit{Aggressive}):

\begin{enumerate}
    \item Stage 2 fine-tuning on the training and development sets of the relevant version of the dataset (\textit{Balanced}, \textit{Gentle} or \textit{Aggressive}) to produce the \textit{BERT-CF} and \textit{BERT-MLM} representation models. BERT-CF is trained following the intervention methodology of Section \ref{subsec:lrm}, while BERT-MLM is trained with standard MLM training as the original BERT model.
    \item Stage 3 supervised-task training for a classifier based on \textit{BERT-O}, \textit{BERT-MLM} or \textit{BERT-CF}, for the relevant downstream task (\textit{Sentiment}, \textit{POMS}, \textit{TC} or \textit{CC} probing).
    \item Test our Stage 3 trained \textit{BERT-O}, \textit{BERT-MLM} and \textit{BERT-CF} based classifiers on the test set of the downstream task. Particularly, the causal and baseline estimators are computed on the test sets.
\end{enumerate}

In all our experiments we utilize the case-sensitive \textit{BERT-base} pre-trained text representation model ($12$ layers, $768$ hidden vector size, $12$ attention heads, $110M$ parameters), trained on the BookCorpus ($800M$ words)~\cite{zhu2015bookcorpus} and Wikipedia ($2,500M$ words) corpora, which is publicly available along with its source code via the Google Research GitHub repository.\footnote{\url{https://github.com/google-research/bert}}

When training \textit{BERT-CF} (Stage 2) we fine-tune all $12$ layers of BERT.  For the downstream task classifier we employ a fully connected layer that receives as input the token representations produced by BERT's top layer, as well as its $CLS$ token. For INLP, which requires a single representation vector as input, we provide the model with the $CLS$ token of the \textit{BERT-O}'s top layer. The output produced by INLP for this vector is then provided to the default classifier of that work, namely logistic regression, in order to compute the $\text{TReATE}(O,INLP)$ estimator.

All our models use cross entropy as their loss function. We employ the ADAM optimization algorithm~\cite{kingma2015adam} with a learning rate of $1e^{-3}$, fuzz factor of $1e^{-8}$ and no weight decay. We developed all our models and experimental pipelines with PyTorch~\cite{paszke2017pytorch}, utilizing and modifying source code from HuggingFace's "Transformers"~\cite{Wolf2019HuggingFacesTS} and PyTorch Lightning~\cite{falcon2019pytorch} GitHub repositories.\footnote{\url{https://github.com/huggingface/transformers}, \url{https://github.com/PyTorchLightning/pytorch-lightning}}

Due to the extensive experimentation pipeline, which resulted in a large total number of experiments over many different combinations of dataset versions and model variations, we chose not to tune our hyper-parameters. Table~\ref{tab:bert_params} details the hyper-parameters used for all our developed models in all experiments.